\pdfoutput=1

\documentclass[11pt]{article}

\usepackage[final]{acl}

\usepackage{booktabs}
\usepackage{times}
\usepackage{latexsym}
\usepackage{multirow}
\usepackage{amsmath}
\usepackage[T1]{fontenc}

\usepackage[utf8]{inputenc}

\usepackage{microtype}

\usepackage{inconsolata}

\usepackage{graphicx}

\title{Are Stereotypes Leading LLMs’ Zero-Shot Stance Detection ?}

\author{
 \textbf{Anthony Dubreuil\textsuperscript{1}},
 \textbf{Antoine Gourru\textsuperscript{1}},
 \textbf{Christine Largeron\textsuperscript{1}},
 \textbf{Amine Trabelsi\textsuperscript{2}}
 \\
 \textsuperscript{1}Laboratoire Hubert Curien, UMR CNRS 5516, Saint-Etienne, France, \\
 \textsuperscript{2}Department of Computer Science, Université de Sherbrooke,  Canada,\\
 \texttt{anthony.dubreuil@etu.univ-st-etienne.fr, antoine.gourru@univ-st-etienne.fr,} \\
 \texttt{ christine.largeron@univ-st-etienne.fr, amine.trabelsi@usherbrooke.ca}\\
 \small{
\textbf{Correspondence:} \href{mailto:antoine.gourru@univ-st-etienne.fr}{antoine.gourru@univ-st-etienne.fr}
 }
}

\begin{document}
\maketitle
\begin{abstract}
Large Language Models inherit stereotypes from their pretraining data, leading to biased behavior toward certain social groups in many Natural Language Processing tasks, such as hateful speech detection or sentiment analysis. Surprisingly, the evaluation of this kind of bias in stance detection methods has been largely overlooked by the community. Stance Detection involves labeling a statement as being against, in favor, or neutral towards a specific target and is among the most sensitive NLP tasks, as it often relates to political leanings. In this paper, we focus on the bias of Large Language Models when performing stance detection in a zero-shot setting. We automatically annotate posts in pre-existing stance detection datasets with two attributes: dialect or vernacular of a specific group and text complexity/readability, to investigate whether these attributes influence the model's stance detection decisions. Our results show that LLMs exhibit significant stereotypes in stance detection tasks, such as incorrectly associating pro-marijuana views with low text complexity and African American dialect with opposition to Donald Trump. 
\end{abstract}

\section{Introduction}

Large Language Models (LLMs) are computational models with billions of parameters, demonstrating exceptional performance across various Natural Language Processing (NLP) tasks. A notable example is ChatGPT, commonly used for question answering and writing assistance. LLMs are not limited to text generation. They also excel in summarization, translation, text classification, and other core NLP functions.

Previous studies indicate that prompt engineering, i.e., optimizing input instructions, can sometimes outperform traditional NLP model tuning for specific tasks \cite{kheiri2023sentimentgpt}. One such task is stance detection, which infers an author's position on a topic based on the text they wrote. Stance detection models typically classify opinions as "Favorable", "Against", or occasionally "Neutral". LLMs have demonstrated strong performance in stance detection, surpassing specialized models \cite{cruickshank2024prompting}.

Nevertheless, despite their advanced capabilities, LLMs exhibit significant biases toward social groups. For example, they may default to assuming a doctor is male and a nurse is female, which can impair task performance \cite{salinas2023im,motoki_more_2024,gallegos2024bias,li2024mitigating}. In stance detection, these biases could result in unfair outcomes, such as associating certain ideologies with specific demographic groups, demonstrating the existence of \emph{stereotypes} in the model's parametric knowledge. Here, we refer to a stereotype as the set of ideas used to describe a person or a social group that is often reducing or false\footnote{\url{https://dictionary.cambridge.org/dictionary/english/stereotype}}. 

Surprisingly, limited research has focused on bias in stance detection, particularly regarding racial and social group biases in LLMs, even though a recent study showed that language models demonstrate gender bias in stance detection \cite{li2024pro}. This gap is especially concerning given the task's sensitivity and its potential real-world impact, such as inferring a social media user's political orientation. Moreover, the scarcity of datasets that integrate both stance information and author attributes significantly limits the ability to study and mitigate bias in this domain. As a consequence, \citet{li2024pro} focus on template-based gender bias (i.e. synthetic data), while our work is the first to leverage demographic linguistic cues on real-life data.

In this work, we aim to address the gap in research regarding bias in zero-shot stance detection with LLMs. Our contributions are as follows: (1) We investigate biases in LLMs' stance detection predictions, focusing on discriminatory decisions based on pre-existing stereotypes embedded in their parametric knowledge, such as associating political stances with a vernacular expression of English or text complexity. We evaluate popular LLMs, including Mistral, Llama, Falcon, Flan, and GPT-3.5, on stance detection tasks and analyze their biases using several fairness metrics. (2) We release enhanced datasets that integrate stance information with sensitive attributes for further research. (3) Our findings reveal significant biases, including the association of certain political and social issues with specific sensitive attributes, emphasizing the need for more equitable stance detection models and better debiasing techniques.


\section{Related Work}

\subsection{Stance Detection}



Stance detection used to be dominated by supervised methods, often enhanced by pre-trained language models \cite{hfconvbilstm}, or unsupervised approaches \cite{sutter-etal-2024-unsupervised}. Recently, modern Language Models were shown to be fast learners, and demonstrate good abilities in zero-shot settings \cite{kocon2023chatgpt}. \citet{cruickshank2024prompting} show that, under the usage of effective prompting methods, LLMs are able to outperform baselines on the stance detection task. Therefore, in the past months, stance detection using LLMs has been largely expanded upon. \citet{wang2024deem} work shows even better results with LLMs, using a new method to inject expert information into the models. 

\subsection{Fairness/Bias of Language Models}

In their stance detection benchmark from 2020, \citet{schiller2021stance} do mention the problem of bias in stance detection models, showing that while it has been a known problem for years, little to no research has been done about it. Language models were shown to be biased by many existing studies \cite{dixon2018measuring,kiritchenko2018examining,leteno2023fair}, i.e. they were shown to demonstrate different behavior with regard to the demographic group associated with the text, mostly gender and race. 
\citet{salinas2023im} show ways to prompt a model to remove its filters, confirming obvious bias against certain groups when the model is not restrained by manually applied constraints. \citet{motoki_more_2024} trick ChatGPT into impersonating humans with certain political opinions, leading to biased responses when the model does not consider itself restrained anymore. Additionally, LLMs were shown by \citet{feng2023pretraining} to be politically oriented.

This work shows that politically skewed pretraining data can propagate biases into LLMs' applications, resulting in unfair predictions, especially in tasks involving social or identity groups. This could lead the language model to inherit some biases or stereotypes that might impact its decision when detecting stances toward political subjects such as those appearing in the commonly used datasets, e.g. Biden, Trump, abortion, gay rights \cite{hasan-ng-2013-stance}.

Surprisingly, the issue of bias in stance detection approaches has received little attention in the literature, possibly due to the scarcity of sensitive attribute annotations within existing datasets. \citet{li2024mitigating} examine the potential influence of the text polarity on the model decision, but also target preference, similarly to \citet{zhang2024relative}. Close to the latter, \citet{yuan2022debiasing} use causal graph modeling and propose to isolate the text's direct effect on stance and to focus on the text-target interaction. 

In this paper, we focus on biases as unfair actions that result more often from stereotypes, i.e. over-generalization or false beliefs toward a certain part of the population, most often social groups such as defined by so-called protected attributes (gender, race, etc.).To date, the work of \citet{li2024pro} is the only one that focuses on social group bias in stance detection algorithms. They demonstrate the existence of gender biases in stance detection based on language models such as BERT, GPT-3.5 and GPT-4 in zero-shot settings, using generated data. No other work proposes to study two important sensitive attributes: African American English vs Standard American English and Text complexity, easily detectable with Flesch score, and their influence on the model decision when producing a stance for a text on politically oriented topics.  

\section{Methodology}

In this section, we provide all the information concerning our protocol. Note that in addition, we make the datasets and code available online\footnote{\url{https://github.com/AntoineGourru/StanceDetectionBiases}}.

\subsection{Enriching Datasets with Sensitive Attributes Annotation}

In this paper, we measure bias 
related to two different sensitive attributes.
None of the existing Stance Detection datasets contain text or 
post-level sensitive attribute annotations. Therefore, we propose to leverage existing datasets and augment them with automatic text annotation for two sensitive attributes. We consider the potential bias of the models regarding African-American English (AAE) text. AAE can be grammatically and syntactically different from Standard American English (SAE),
serving as a proxy for linguistic and sociocultural group membership. Importantly, note that as stated in \cite{blodgett-etal-2016-demographic}, ``Not all
African-Americans speak AAE, and not all speakers
of AAE are African-American''. AAE/SAE is used here as a linguistic marker, not as a deterministic racial classifier, and it represents perceived sociocultural identity, which is interpreted by LLMs as a social signal, a central point of our bias hypothesis.
Second, we consider the bias towards text complexity/readability. We use the Flesch-Kincaid score \cite{kincaid1975derivation}, which is a test for the readability of a text or sentence. Our aim is to assess whether models implicitly rely on text complexity to make biased assumptions. Bias in LLMs related to text complexity is especially concerning, as recent work \cite{ahmed2022flesch} found correlations between readability and socio-economic status on social media. In the following section, we detail the methods we used to enrich the existing datasets with these sensitive attributes.

\subsubsection{African vs Standard American English} 

To infer the nature of the language, we propose to leverage the model\footnote{\url{https://github.com/slanglab/twitteraae}} proposed by \citet{blodgett-etal-2016-demographic} as was done to build the MOJI dataset. This model takes a text as input and returns a probability for four possible forms of English, labeled as "African-American", "Hispanic", "Asian", and "Standard". We label every text with the category with the highest probability. In our study, we focus on "African-American" and "Standard American" (SAE). 

\begin{quote}\itshape
"Okay then, I'm on it!!! And remember folks, Greg Gutfeld says he's never met a Biden supporter. \#ImABidenSupporter!"
\end{quote}

\begin{center}
\footnotesize
Example of a text from the PStance dataset labeled as SAE
\end{center}

\begin{quote}\itshape
"Nope that's NOT true we would be respected around the world with @JoeBiden we suffered a recession under @BarackObama guess wat he got us out of that with u only DOWN"
\end{quote}

\begin{center}
\footnotesize
Example of a text from the PStance dataset labeled as AAE
\end{center}

\begin{figure}
\centering
\includegraphics[width=\linewidth]{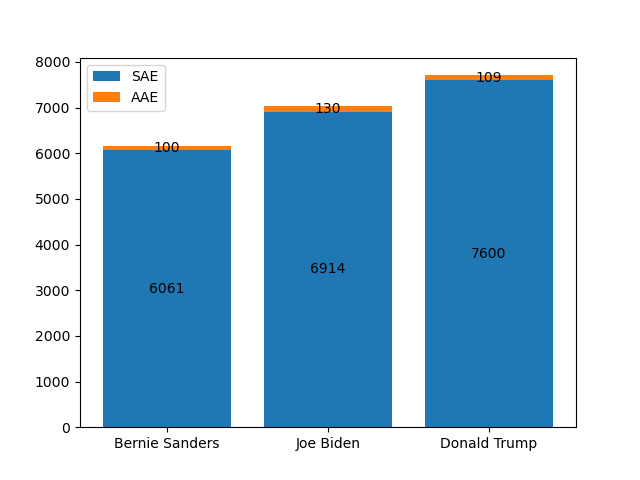}
\caption{Proportion of SAE and AAE tweets in the PStance dataset for each political figure}
\label{method1_proportion}
\end{figure}

\subsubsection{Text Complexity}

To measure the text complexity of a given text, we use the Flesch–Kincaid readability test \cite{kincaid1975derivation}. This test measures the readability of a text by evaluating the average sentence length and the average number of syllables per word. The resulting score corresponds to a reading ease scale, where higher scores indicate easier readability. This approach has been widely used in readability research and serves as a reliable indicator of the text's complexity. The Flesch-Kincaid score is computed as follows: 

 \begin{equation}
 206.835 - 1.015 \times \frac{W}{Se} - 84.6 \times \frac{Sy}{W}
 \end{equation}
 with $Se$ the number of sentences in the text, $W$ the total number of words and $Sy$ the total number of syllables. 

We discretize this score in four groups following previous works: Easy (or low complexity), Medium, Difficult readability and Very Difficult readability (or very high complexity, see Table~\ref{tab:fkd} for details).

\begin{table}[h!]
\centering
\begin{tabular}{|cc|}
\hline
\textbf{Flesch Score} & \textbf{Readability} \\
\hline
$ \ge 80 $ & Easy \\
$ \ge 60, < 80 $ & Medium \\
$ \ge 30, < 60 $ & Difficult \\
$ < 30 $ &Very difficult \\
\hline
\end{tabular}
\caption{Discretization of Flesch-Kincaid Score}
\label{tab:fkd}
\end{table}

We hypothesized that the complexity of a text, measured by the F-K readability tests, could potentially affect the model’s assumptions about the writer’s writing skills. In other words, a high or low language complexity (the quality of writing) of a text might result in biased decisions about its stance.

\begin{quote}\itshape
"I believe that they should be able to because it is their right. Just like we have the right to marry one another they should be able to. How about this put yourself in their shoes how would you like it if you were in love with the same sex and you 2 decide to get married but you couldn't then what? You would be pretty mad wouldn't you?. I know that I would. So to me I think they should be able to get married. "
\end{quote}

\begin{center}
\footnotesize
Example of text from SCD labeled ``Low text complexity''
\end{center}

\begin{quote}\itshape
"To say that two men or two women necessarily can't raise a child as well as a one-man-one-woman couple is sexist and inaccurate. We all know there are some heterosexual couples who are clearly unqualified to raise children. Restrictions on adoption should depend on the individual circumstances of the adopting family, not on generalized statements about the differing parenting styles of men and women."
\end{quote}

\begin{center}
\footnotesize
Example of text from SCD labeled ``Very high text complexity''
\end{center}

\begin{figure}[t]
\centering
\includegraphics[width=\linewidth]{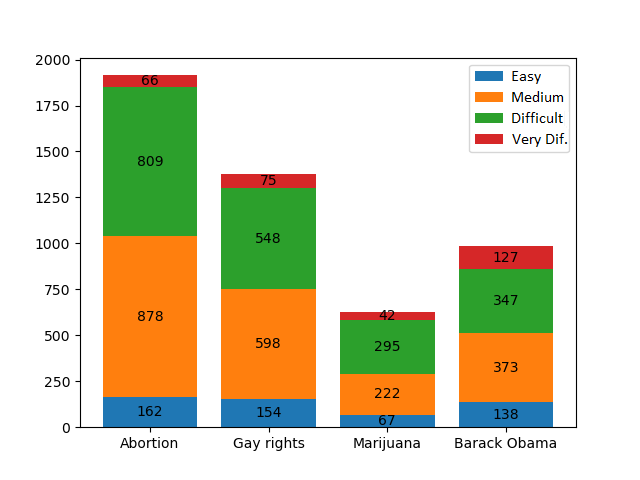}
\caption{Proportion of readability classes in the SCD dataset for each topic}
\label{method2_proportion}
\end{figure}

\subsection{Datasets Used}

For this study, we use existing stance detection datasets for which we create sensitive attributes using the aforementioned methods. We use one dataset for the stereotypical bias toward language variety (SAE vs AAE) and two datasets for the bias toward text complexity.

\begin{figure}[t]
\centering
\includegraphics[width=\linewidth]{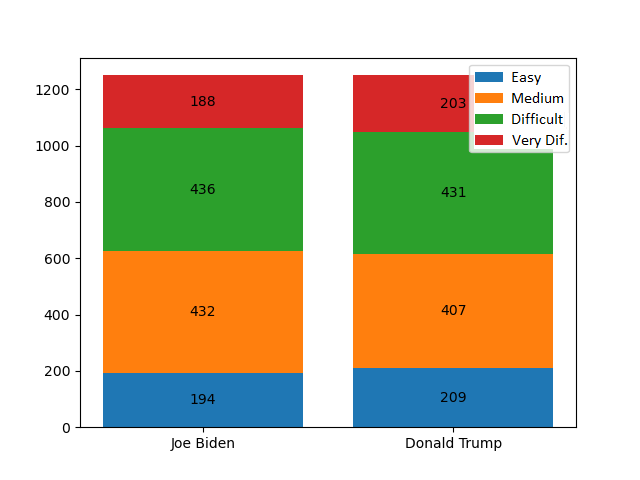}
\caption{Proportion of readability classes in the KE-MLM dataset for each political figure}
\label{method2_kemlm_proportion}
\end{figure}

For the language varieties experiment, we use the PStance dataset \cite{li-etal-2021-p}, a stance detection dataset composed of a large number of posts retrieved from X (formerly Twitter) in the political domain. Specifically, this dataset focuses on three American political figures: Bernie Sanders, Joe Biden and Donald Trump.

After running our sensitive attribute annotation protocol on this dataset, a clear imbalance was shown, with a large majority of the dataset being labeled as SAE tweets, and only a small portion of the dataset being labeled as AAE (see Figure~\ref{method1_proportion}). However, we deemed the numbers sufficient and went ahead with the experiment. For the experiments, we balance the dataset by downsampling the majority group, so that there are as many AAE tweets as SAE tweets in our study, and the same proportion of favorable tweets in both groups.

For the text complexity experiment, we use the SCD Dataset \cite{hasan-ng-2013-stance}, which consists of posts taken from the CreateDebate website. These posts are part of debates about four themes: Abortion, Gay rights, Marijuana and Barack Obama. Since this dataset is sourced from a debating website and consists of long texts, the Flesch-Kincaid reading ease test is more applicable. The proportion of texts for each readability class in the SCD dataset after annotation can be seen in Figure~\ref{method2_proportion}. Additionally, we use the KE-MLM dataset \cite{kawintiranon-singh-2021-knowledge}, another dataset about two political figures, Donald Trump and Joe Biden, which contained substantial numbers of tweets from all four text complexity classes, making it usable to evaluate models' bias. The proportion of tweets for each readability class in the KE-MLM dataset after annotation can be seen in Figure~\ref{method2_kemlm_proportion}.

The choice to use the PStance, SCD, and KE-MLM datasets in different contexts was driven by the specific characteristics of each dataset and the requirements of our experiments. The PStance dataset was ideal for language variety experiments due to its focus on diverse linguistic expressions across various stances. For text complexity experiments, the SCD dataset was initially selected because it contains longer texts, making it more suitable for applying the Flesch-Kincaid readability test effectively. Later, we incorporated the KE-MLM dataset for text complexity experiments to explore whether similar patterns observed in longer texts could also emerge in shorter, more dynamic texts like tweets. The PStance dataset, however, did not yield meaningful results for the text complexity experiments due to a large over representation of medium- and low-complexity levels, rendering it unsuitable for our analysis. 
In contrast, while the KE-MLM dataset also consists of tweets, the Flesch-Kincaid test provided a more balanced group distribution (see Figure 2). However, the SCD and KE-MLM datasets included an insignificant proportion of AAE texts, making our study on language varieties inapplicable to them. 

For all datasets, we balance the data to ensure an equal proportion of favorable and unfavorable posts for each class. Although this results in smaller datasets, it mitigates the potential bias caused by class imbalance. The initial statistics for each dataset are provided in the Appendix. 

\subsection{Language Model and Prompting}

As we evaluate zero-shot stance detection, we use one closed model, GPT-3.5-turbo-0125, and four open models, Llama3-8B-Instruct, Mistral-7B-Instruct-v0.2, Falcon-7b-instruct and FLAN-T5-large. We provide URLs in the Appendix. 

Several prompting methods can be used to perform stance detection with LLMs. Among those described by \citet{cruickshank2024prompting}, we employ the Context Analyze and Zero-shot Chain-of-Thought methods, as both demonstrated superior results with Mistral compared to other approaches. Since both methods yielded similar outcomes in preliminary experiments, we opted for the Context Analyze method due to its significantly faster performance compared to Zero-Shot Chain-of-Thought. Following the Context Analyze method we use the prompt:

\begin{quote}\itshape
Stance classification is the task of determining the expressed or implied opinion, or stance, of a statement toward a certain, specified target.\\
Analyze the following social media statement and determine its stance towards the provided [target]. Respond with a single word: FAVOR or AGAINST. Only return the stance as a single word, and no other text.\newline
[target]: \textbf{TARGET}\\
Statement: \textbf{TEXT}\\
\label{prompt}
\end{quote}

with:
\begin{itemize}
\item \textit{\textbf{TARGET}} replaced with the subject we want to detect the stance about
\item \textit{\textbf{TEXT}} replaced with the full  text
\end{itemize}

\subsection{Measures}  
  
As done in previous works, we rely on weighted F1 as a measure of performance for the (binary) stance detection evaluation, 1 being the best score. With regard to fairness, we rely on Equal Opportunity (EO), and extend to Demographic Parity and Predictive Parity in the Appendix \ref{App:metrics} \cite{alves2023survey}. In the sequel, $ y $ denotes the stance label,  $ \hat{y} $ the prediction made by the model, $ s $  a sensitive attribute, taking values corresponding to different groups ($a$ and $\bar{a}$). Equal Opportunity (EO) is defined by:

\begin{equation}
\begin{split}
  EO &= p(\hat{y} = 1 | y = 1, s = a) \\
  &- p(\hat{y} = 1 | y = 1, s = \bar{a})  
\end{split}
\end{equation}

EO ranges from $-1$ to $1$, with $0$ being the fairer result, $-1$ meaning that group $a$ is discriminated by the model (less likely to predict 1 for examples labeled 1 and with sensitive attribute value $a$) and $1$ meaning that group $a$ is privileged by the model (more likely to predict 1 for examples labeled 1 and with sensitive attribute value $a$). Equal Opportunity \cite{hardt2016equality} allows us to compare the probability of labeling a text 1 with property $a$ to the probability of labeling 1 a text without property $a$, knowing that the true label of the  text is 1. In our experiment, we present EO in both ways: with label 1 corresponding to "favor" and then to "against".

We also provide an aggregated version of EO, by computing the average of absolute values of EO for each class, dataset and stance, allowing us to compute the overall language model bias for the considered sensitive attributes. 

\section{Results}

\subsection{Stance Detection Capabilities}

The weighted F1-score of each model on each dataset can be found in Table~\ref{f1_scores}. To put these results into perspective, we also provide the percentage of "Neutral" predictions made by each model on each dataset in the Appendix (F1-score is computed only on the favor and against stances).

The results from Table~\ref{f1_scores} indicate a clear performance hierarchy among the evaluated models. Falcon is the least effective model, demonstrating the lowest performance. In contrast, Llama achieves good results in terms of F1-score. However, Llama generates a high number of neutral predictions, which raises concerns about its overall reliability and effectiveness for this task. This tendency towards neutrality suggests that Llama may struggle to predict stance in zero-shot settings, limiting its practical application.

Flan shows above-average capabilities, indicating it is a strong contender for stance detection tasks. Its performance is consistently reliable, making it a dependable choice for researchers and practitioners. However, Flan does not outperform the top models, Mistral and GPT-3.5, which demonstrate superior performance in the task.

Mistral and GPT-3.5 emerge as the best models for stance detection. Among these, GPT-3.5 is particularly noteworthy for its significantly lower number of neutral predictions. This characteristic indicates that GPT-3.5 is more decisive and confident in its classifications, making it highly effective for tasks requiring clear and definitive stances.

\begin{table}
    \centering
    \footnotesize
    \begin{tabular}{ccccccc}
    \toprule
         & Mistral & Llama & Falcon & Flan & GPT \\\midrule
         PStance & \textbf{0.804} & 0.711 & 0.477 & 0.693 & 0.787 \\\midrule
         SCD & 0.637 & 0.617 & 0.513 & 0.591 & \textbf{0.685} \\
         KE-MLM & 0.671 & 0.639 & 0.494 & 0.623 & \textbf{0.695} \\
         \bottomrule
    \end{tabular}
    \caption{Weighted F1 for each dataset and LLM}
    \label{f1_scores}
\end{table}
\subsection{Biases of LLMs}

\begin{figure}[t]
    \centering 
    \includegraphics[width=\linewidth]{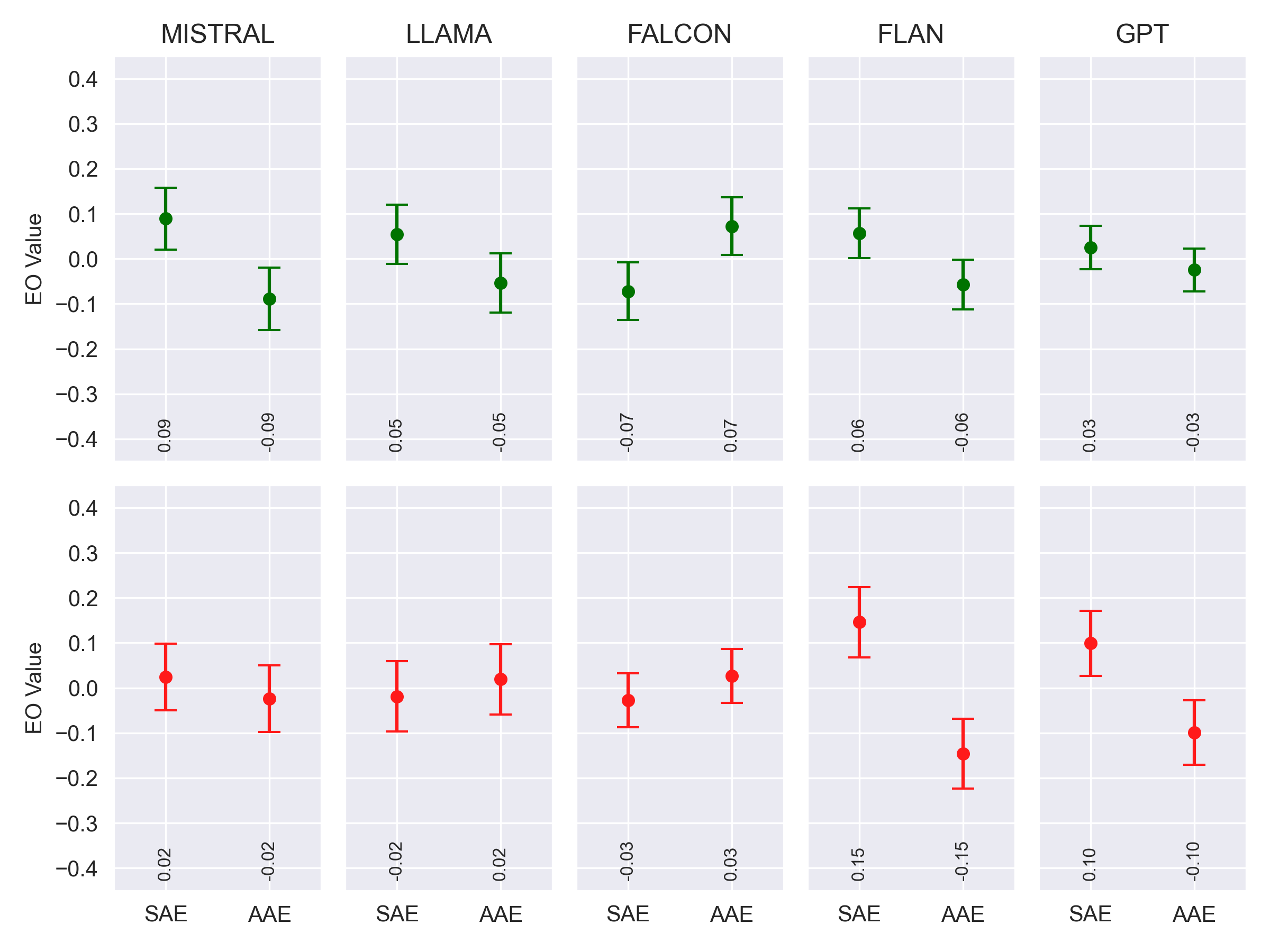}
    \caption{Equality of Opportunity on Joe Biden on the PStance dataset. SAE stands Standard American English, AAE for African American English. In green, EO for the label "favor", in red for ``against''.}
    \label{eo_p_biden}
\end{figure}

\begin{figure}[t]
    \centering
    \includegraphics[width=\linewidth]{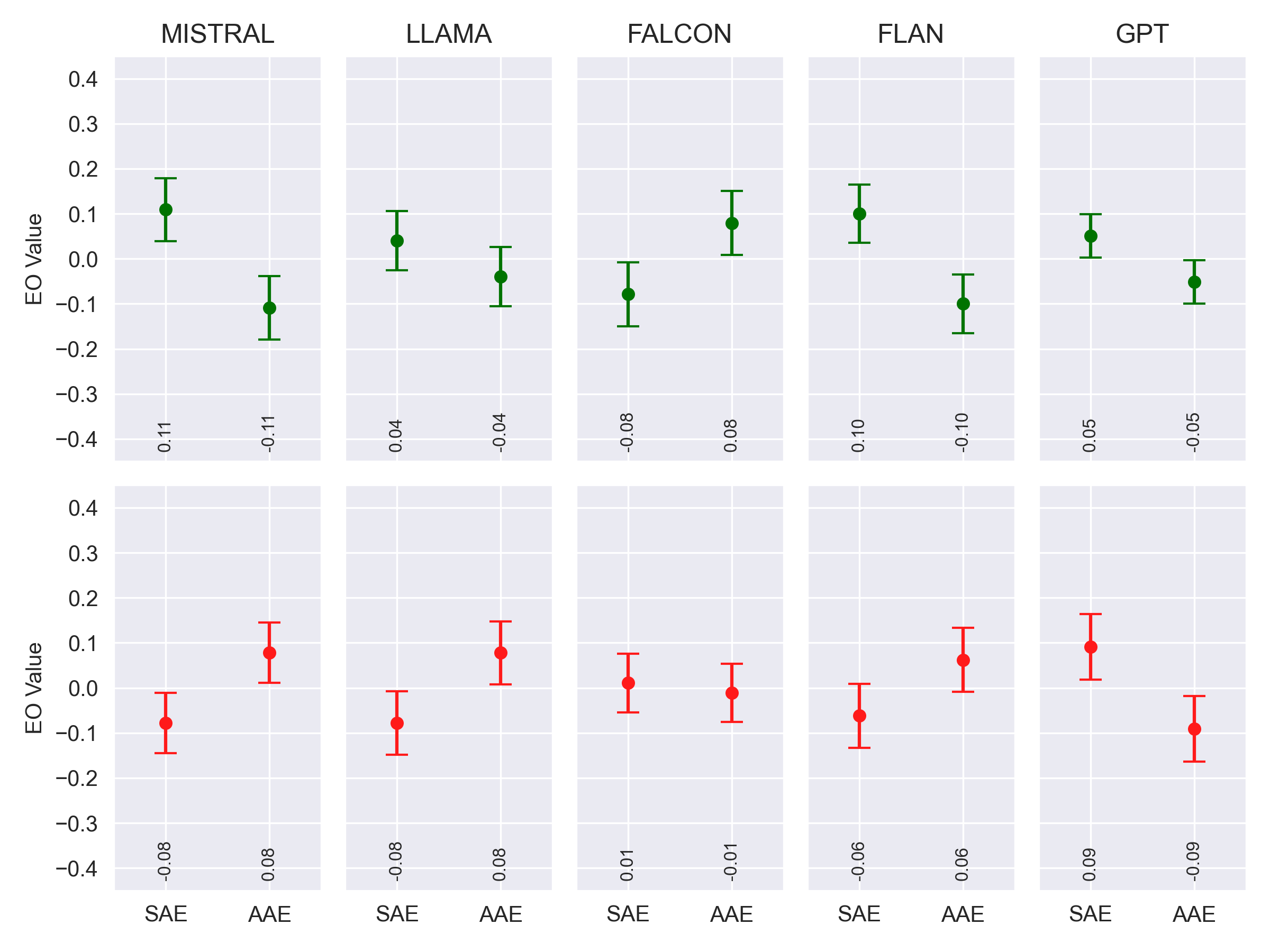}
    \caption{Equality of Opportunity on Bernie Sanders on the PStance dataset. SAE stands Standard American English, AAE for African American English. In green, EO for the label "favor", in red for ``against''.}
    \label{eo_p_sanders}
\end{figure}

\begin{figure}[t]
    \centering
    \includegraphics[width=\linewidth]{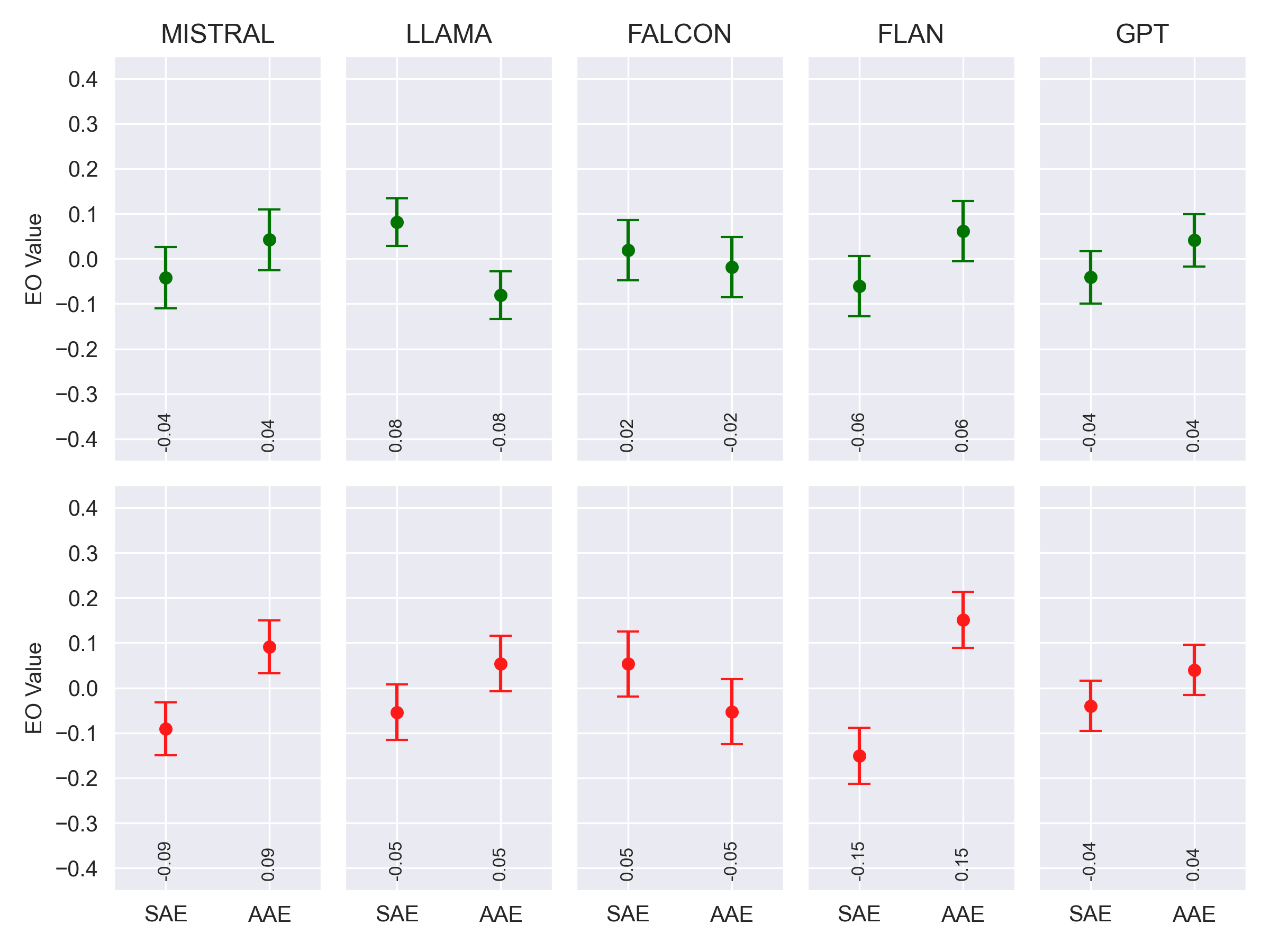}
    \caption{Equality of Opportunity on Donald Trump on the PStance dataset. SAE stands Standard American English, AAE for African American English.In green, EO for the label "favor", in red for ``against''.}
    \label{eo_p_trump}
\end{figure}

\begin{figure}[t]
    \centering
    \includegraphics[width=\linewidth]{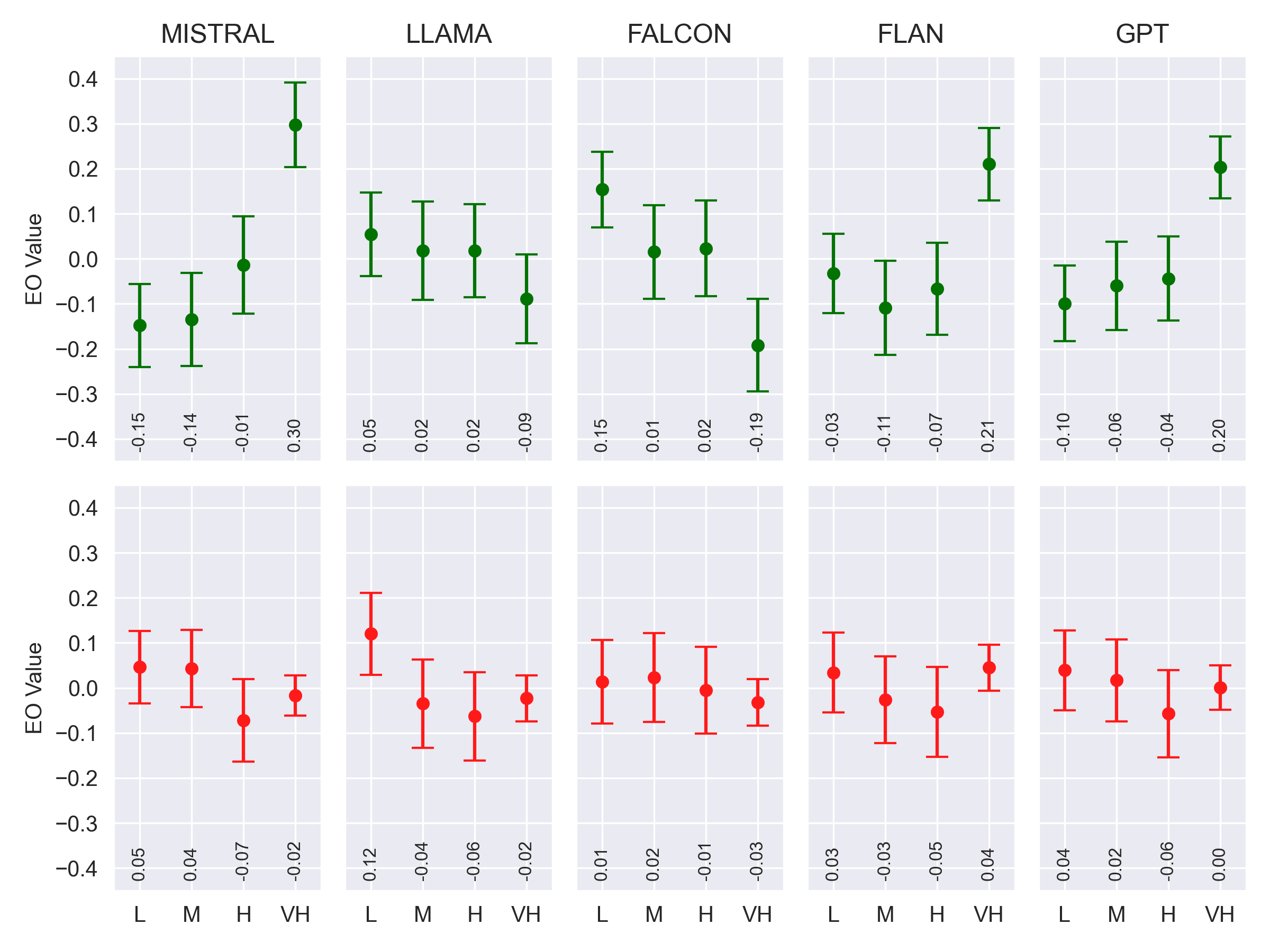}
    \caption{Equality of Opportunity on Barack Obama on the SCD dataset. L, M, H, and VH stand for Low - Medium - High and Very High Text Complexity. In green, EO for the label "favor", in red for ``against''.}
    \label{eo_obama}
\end{figure}

\begin{figure}[t]
    \centering
    \includegraphics[width=\linewidth]{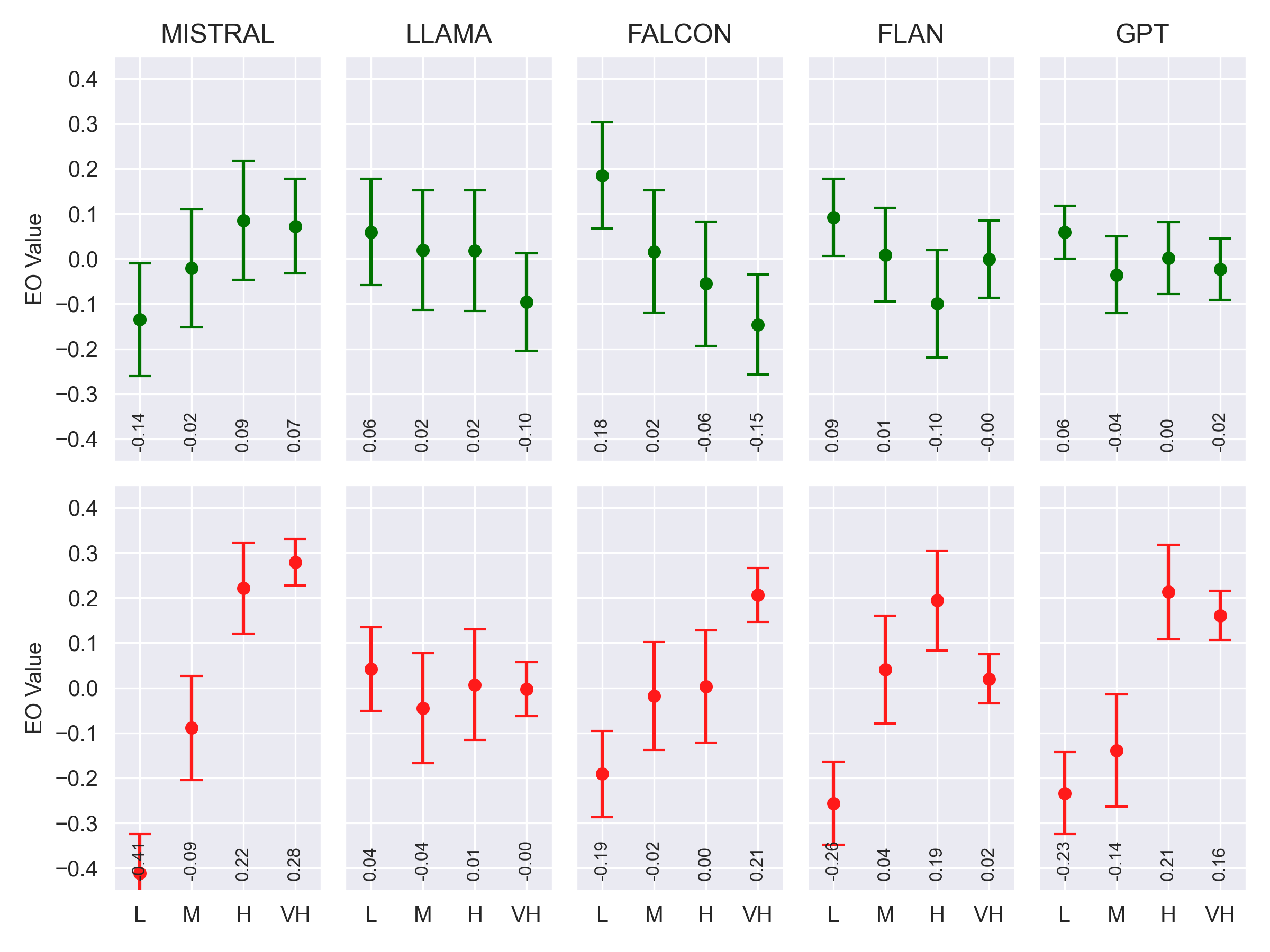}
    \caption{Equality of Opportunity on Marijuana on the SCD dataset. L, M, H, and VH stand for Low - Medium - High and Very High Text Complexity, respectively. In green, EO for the label "favor", in red for ``against''.}
    \label{eo_marijuana}
\end{figure}

\begin{figure}[t]
    \centering
    \includegraphics[width=\linewidth]{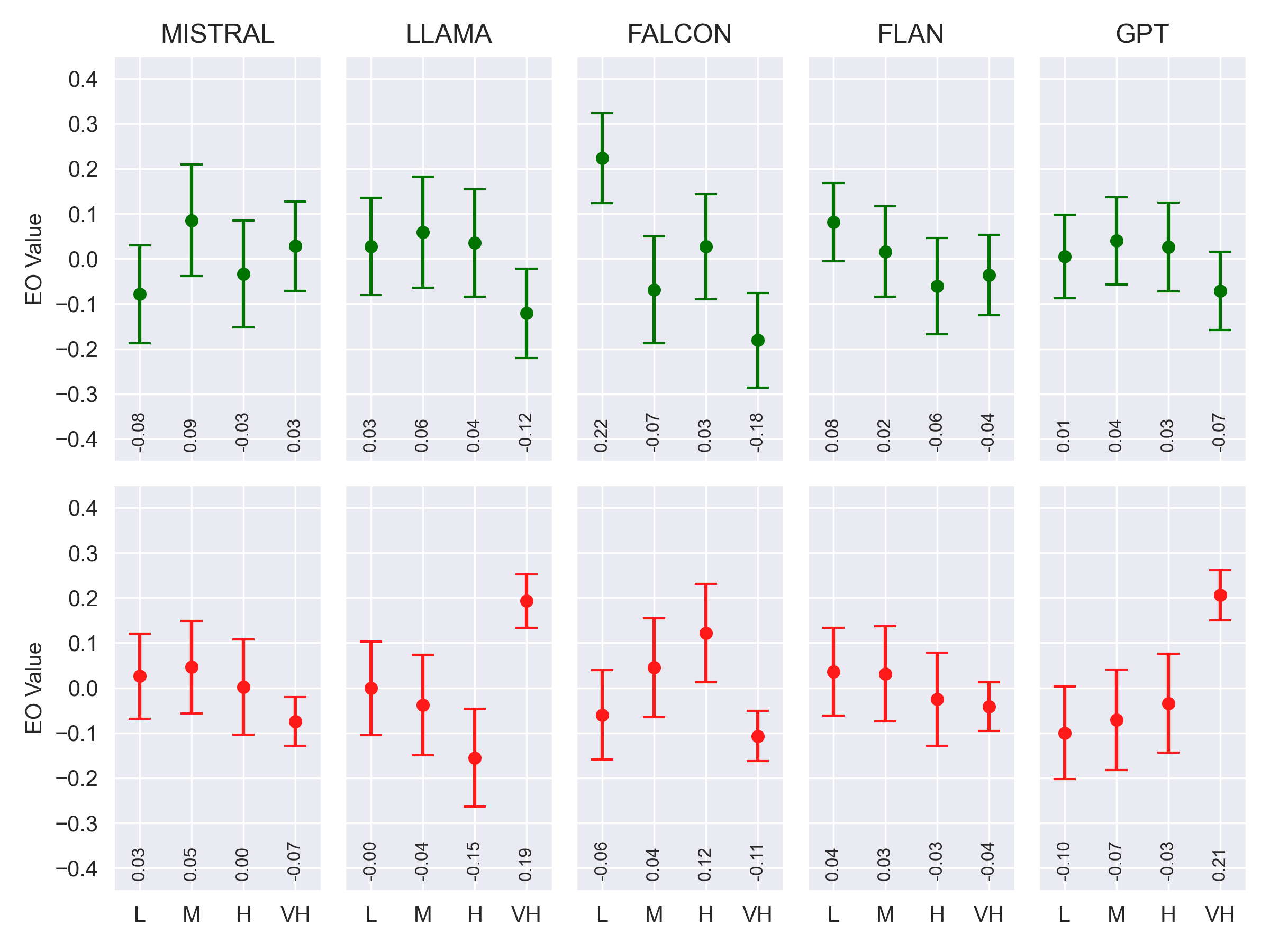}
    \caption{Equality of Opportunity on gay rights on the SCD dataset. L, M, H, and VH stand for Low - Medium - High and Very High Text Complexity, respectively. In green, EO for the label "favor", in red for ``against''.}
    \label{eo_gay}
\end{figure}

\begin{figure}[t]
    \centering
    \includegraphics[width=\linewidth]{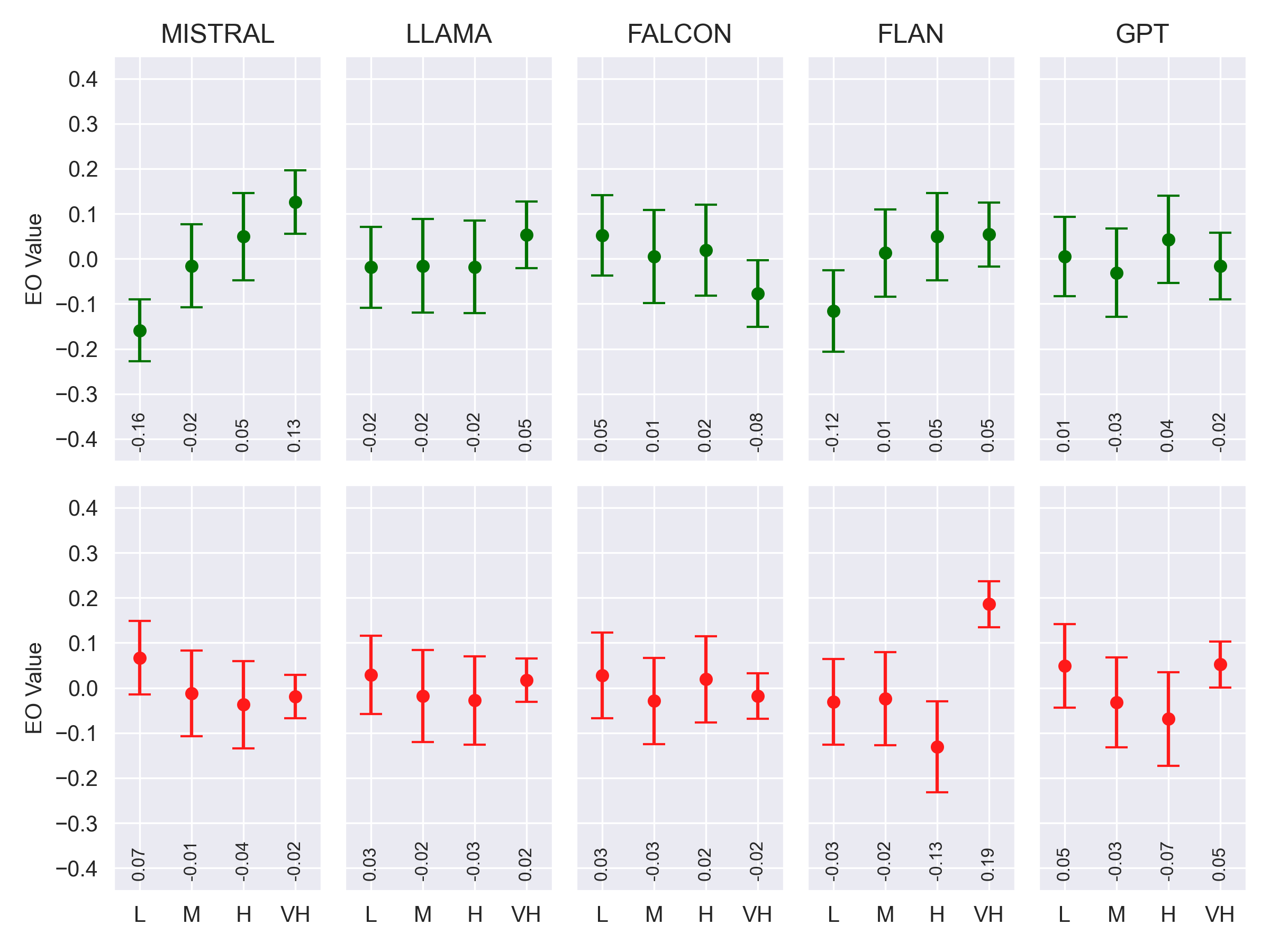}
    \caption{Equality of Opportunity on abortion  on the SCD dataset. L, M, H, and VH stand for Low - Medium - High and Very High Text Complexity, respectively. In green, EO for the label "favor", in red for ``against''.}
    \label{eo_abortion}
\end{figure}

\begin{figure}[t]
    \centering
    \includegraphics[width=\linewidth]{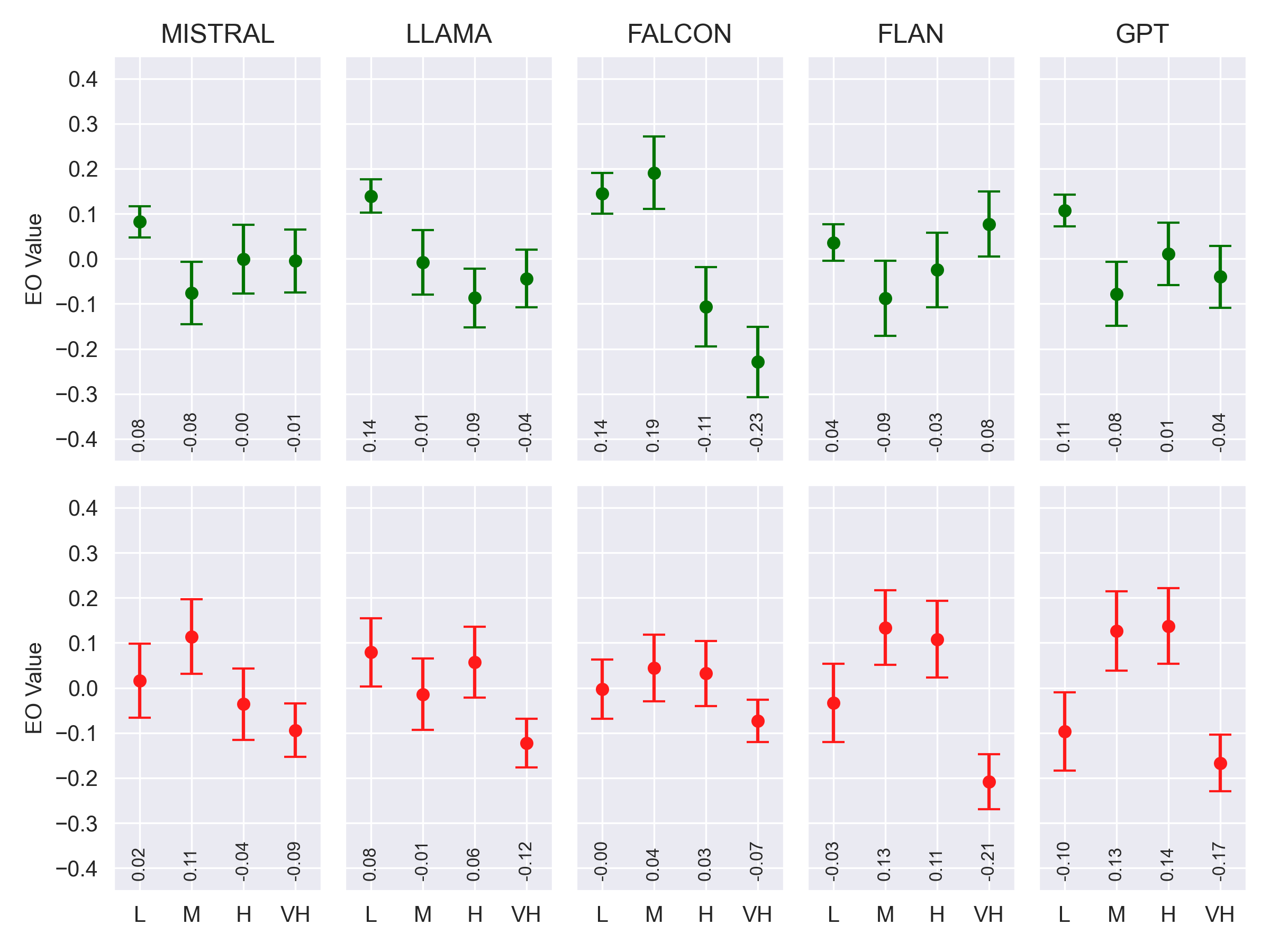}
    \caption{Equality of Opportunity on Joe Biden on the KE-MLM dataset. L, M, H, and VH stand for Low - Medium - High and Very High Text Complexity. In green, EO for the label "favor", in red for ``against''.}
    \label{eo_biden}
\end{figure}

\begin{figure}[t]
    \centering
    \includegraphics[width=\linewidth]{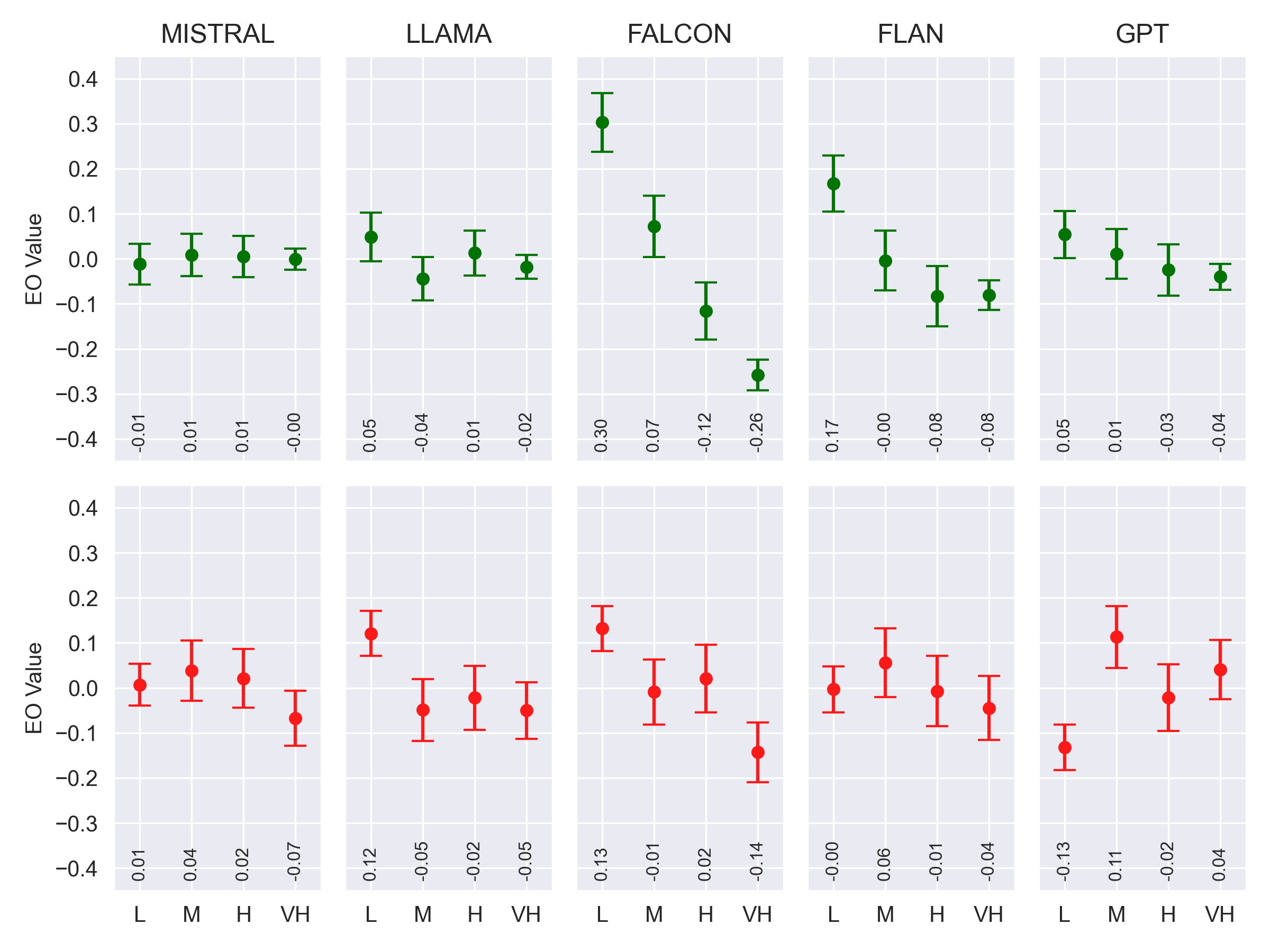}
    \caption{Equality of Opportunity on Donald Trump on the KE-MLM dataset. L, M, H, and VH stand for Low - Medium - High and Very High Text Complexity. In green, EO for the label "favor", in red for ``against''.}
    \label{eo_trump}
\end{figure}

All fairness results have been computed by averaging the metrics on 1000 balanced samples randomly taken from the dataset. Each sample contains an equal number of texts for each class, and as many favorable and unfavorable texts in each class. The same samples have been used for the 5 models. We provide the average EO (with standard deviation) per group for each class and target. More precisely, a dot indicates the average result (the value is given on the bottom) and the whiskers represent standard deviation (mean -/+ sd). On top of each EO plot, we provide results in green when "favor" is considered as label 1, and on the bottom in red when ``against'' is considered as label 1. This allows to show the bias toward each target in a single plot. For instance, in Figure~\ref{eo_abortion} related to abortion, with EO values around 0, LLMs demonstrate limited biases w.r.t. the text complexity. 

\paragraph{On AAE vs SAE bias} Figures~\ref{eo_p_biden},~\ref{eo_p_sanders} and ~\ref{eo_p_trump} present the results on the PStance dataset. Surprisingly, results seem to show very little bias based on African American English. The only low magnitude biases we observe are the following: the FLAN model seems to associate more easily SAE as being against Biden than AAE. Similarly, FLAN seems to associate AAE more easily as being against Trump than SAE. 

\paragraph{Stereotype 1: 
Low complexity text
means
in favor of Marijuana - 
Complex text means
against.} 
By examining Figure~\ref{eo_marijuana}, LLMs demonstrate clear biases for all models except Llama and to a lesser extent Falcon on the "against marijuana" target (bottom plots), with values reaching -0.4  for Mistral. The models show a lower probability of predicting low complexity texts as being against marijuana, compared to other groups. In contrast, the models are more likely to predict high complexity texts as being against marijuana. This trend is further supported by additional fairness metrics (see Appendix \ref{sec:appendix}), demonstrating that LLMs tend to associate a highly complex text with opposition to marijuana and a lower complexity
with support toward it.

\paragraph{Stereotype 2: 
Complex text is expressing support to
Obama.} By examining Figure~\ref{eo_obama}, we observe an interesting pattern concerning the target "Barack Obama" on the SCD dataset. All models, except Llama, exhibit biases. Notably, Falcon shows an opposite bias compared to Mistral, Flan, and GPT-3.5 when predicting the label "favor". The latter models tend to predict that high complexity texts favor Obama, while Falcon is more likely to predict that low complexity texts are in favor, and high complexity texts are against.

\paragraph{Stereotype 3: 
Highly complex text is not expressing a stance against Biden.} Analyzing Figure~\ref{eo_biden}, we observe that all models show a bias toward predicting that very high complexity texts are less likely to oppose Biden. This bias is most pronounced in Flan and GPT-3.5, where the probability of classifying highly complex texts as being against Biden is much less than for other complexities. Notably, Falcon once again exhibits an opposite bias for very high complexity and "favor" (shown in green). 

\paragraph{Stereotype 4: GPT-3.5 and Llama believe 
highly complex text expresses a stance
against gay rights.} In Figure~\ref{eo_gay}, a significant bias is observed in GPT-3.5 and Llama predictions related to the stance of high complexity texts on gay rights. These models disproportionately predict that high complexity texts are against gay rights compared to low complexity texts. 

\paragraph{Stereotype 5: Falcon associates low complexity with partisanship, high 
complexity
with opposition} Interestingly, Falcon demonstrates a similar pattern for all three political figures (Figure~\ref{eo_biden}, \ref{eo_obama}, \ref{eo_trump}) in the KE-MLM dataset: in green, it assigns a much higher probability for texts with low complexity to be in favor of the politician than for those with high or very high complexity, regardless of political party. This could suggest that it is more likely to associate simpler text with partisanship and complex posts with opposition.

\begin{table}[]
\small
\begin{tabular}{cccccc}
\toprule
           & Mistral & Llama & Falcon & Flan & GPT  \\ \midrule
Complexity & 0.12    & 0.07  & 0.20   & 0.12 & 0.08 \\
A/SAE & 0.09    & 0.08  & 0.07   & 0.08 & 0.04\\
\bottomrule
\end{tabular}
\centering
\caption{Average absolute value of EO for each model and demographic group.}
\label{eo_average}
\end{table}

\paragraph{Comparison between Language Models} Table~\ref{eo_average} provides the average EO for each model and
studied attribute.
Falcon demonstrates the maximum bias overall, followed by Mistral, Flan, Llama and GPT-3.5. As models are based on a similar architecture,
this difference might stem from either the pre-training corpus or the instruction data, as we used instruction-tuned 
open 
models.

\section{Discussion and Conclusion}

Our study revealed that Large Language Models consistently (across topics and models) exhibit significant biases in zero-shot stance detection, with stereotypes influencing their predictions based on English dialect and text complexity. 
This aligns with the work of \citet{feng2023pretraining}, which traces political and social biases. We hypothesize that differences in bias between models are likely due to variations in their training data composition and instruction tuning strategies.
These biases, which manifest in politically sensitive contexts, highlight the need for closer scrutiny of LLM behavior, particularly in zero-shot settings. Our findings emphasize the importance of developing more robust and equitable stance detection models to mitigate the harmful impacts of such biases. This could be achieved using fairness-aware prompting or calibration, such as discussed in \citet{li2024mitigating},  to reduce bias in predictions or by causal modeling, like counterfactual inference \cite{yuan2022debiasing}, to isolate the contribution of sensitive attributes. Finally, note that our protocol could be generalized to any other group categorization, e.g. gender. This being said, a promising line of research would be to combine static and LLM-based metrics for automatic group categorization. 

\section*{Limitations}


The methods used to create and balance our datasets resulted in relatively small sample sizes. While these sizes are sufficient to demonstrate bias, a study with a larger dataset would be beneficial.

Additionally, the Mistral and Llama versions used in this study have a limited number of parameters. While this allows their use, larger model variants may perform better on the stance detection task and reveal additional biases. 

An important note: AAE is used in this work as a linguistic marker of a dialect, not as a deterministic racial classifier, which could be interpreted by LLMs as a social signal, a central point of our bias hypothesis. As cited previously, "Not all African Americans speak AAE, and not all AAE speakers are African American" \cite{blodgett-etal-2016-demographic}. 

\section*{Acknowledgments}

This research was partially supported by the Natural Sciences and Engineering Research Council of Canada (NSERC) under the Grant No. RGPIN-2022-04789, and partially funded by the French National Research Agency (ANR) in the context of the Diké and FAMOUS projects. 

\bibliography{custom}

\newpage

\appendix

\section{Additional Implementation details}

\subsection{Language Models}

We use the following url/sources for each models: GPT-3.5-turbo-0125 \url{https://platform.openai.com/docs/models/o1}, Llama3-8B-Instruct \url{https://huggingface.co/meta-llama/Meta-Llama-3-8B-Instruct}, Mistral-7B-Instruct-v0.2 \url{https://huggingface.co/mistralai/Mistral-7B-Instruct-v0.2}, Falcon-7b-instruct \url{https://huggingface.co/tiiuae/falcon-7b-instruct} and FLAN-T5-large \url{https://huggingface.co/docs/transformers/model_doc/flan-t5}. 

\subsection{Resampling Statistics}

We provide in Tables \ref{tab:balpst}, \ref{tab:balscd} and \ref{tab:balkm} the resampling statistics for our experiments. 

\begin{table}[ht]
\centering
\footnotesize
\begin{tabular}{lcc}
\hline
 & \textbf{Unbalanced} & \textbf{Balanced} \\
\hline
SAE & 20,575 & 339 \\
AAE & 339 & 339 \\
\hline
\end{tabular}
\caption{PStance dataset balanced/unbalanced distribution}
\label{tab:balpst}
\end{table}

\begin{table}[ht]
\centering
\footnotesize
\begin{tabular}{lcc}
\hline
 Complexity & \textbf{Unbalanced} & \textbf{ Balanced} \\
\hline
Low & 521 & 262  \\
Medium & 2071 & 262  \\
High & 1999 & 262  \\
Very high & 310 & 262  \\
\hline
\end{tabular}
\caption{SCD dataset balanced/unbalanced distribution}
\label{tab:balscd}
\end{table}

\begin{table}[ht]
\centering
\footnotesize
\begin{tabular}{lcc}
\hline
 Complexity & \textbf{Unbalanced} & \textbf{Balanced} \\
\hline
Low & 403 & 160 \\
Medium &  839 & 160 \\
High & 867 & 160 \\
Very high & 391 & 160 \\
\hline
\end{tabular}
\caption{KE-MLM dataset balanced/unbalanced distribution}
\label{tab:balkm}
\end{table}

\begin{table}[h]
    \centering
    \begin{tabular}{lcc}
        \toprule
        \textbf{File} & \textbf{Size} & \textbf{Ratio Favor} \\
        \midrule
        KE-MLM & 1603 & 0.45 \\
        Donald Trump & 840 & 0.41 \\
        Joe Biden & 763 & 0.50 \\
        \midrule
        PStance & 20914 & 0.48 \\
        Bernie Sanders & 6161 & 0.56 \\
        Donald Trump & 7709 & 0.46 \\
        Joe Biden & 7044 & 0.44 \\
        \midrule
        SCD & 4901 & 0.59 \\
        Abortion & 1915 & 0.56 \\
        Barack Obama & 985 & 0.53 \\
        Gay rights & 1375 & 0.64 \\
        Marijuana & 626 & 0.71 \\
        \bottomrule
    \end{tabular}
    \caption{Dataset Size and Ratio}
    \label{tab:data}
\end{table}

\section{Neutral Predictions}

Some LMs fail to follow the prompt and output neutral predictions. We provide the statistics in table~\ref{neutrals}. "Neutral" refers to instances where the model did not return "FAVOR" or "AGAINST" as instructed in the prompt

\begin{table}
    \centering
    \footnotesize
    \begin{tabular}{cccccc}
        \toprule
         & Mistral & Llama & Falcon & Flan & GPT \\\midrule
         PStance & 21.73 & 61.27 & 12.69 & \textbf{0.01} & 0.04 \\\midrule
         SCD & 15.49 & 37.03 & 5.94 & \textbf{0.00} & 0.14 \\
         KE-MLM & 65.96 & 63.84 & 20.00 & \textbf{0.00} & 0.12 \\
         \bottomrule
    \end{tabular}
    \caption{Percentage of neutral predictions made by each model on the datasets. "Neutral" refers to instances where the model did not return "FAVOR" or "AGAINST" as instructed in the prompt}
    \label{neutrals}
\end{table}

\section{Additional Fairness Metrics}
\label{App:metrics}

Disparate Impact measures the probability for a text written by an author belonging to the modality $a$ to be classified as in favor of the target, compared to a text written by someone else.

\begin{equation}
\begin{split}
  DI &= p(\hat{y} = 1 | S = a)  \\
  & - p(\hat{y} = 1 | S = \bar{a})
\end{split}
\end{equation}

Predictive Parity measures the probability for a text written by an author belonging to the modality $a$ to be in favor of the target, compared to a text written by someone else, knowing that the text was classified as in favor of the target by the model.

\begin{equation}
\begin{split}
  PP &= p(y = 1 | \hat{y} = 1, S = a) \\
  &- p(y = 1 | \hat{y} = 1, S = \bar{a})  
\end{split}
\end{equation}

DI and PP range from -1 to 1, with 0 being the fairer result, -1 meaning that the modality a is discriminated by the model and 1 meaning that the modality a is privileged by the model.

\section{Complete results}
\label{sec:appendix}

\begin{figure}
    \centering
    \includegraphics[width=\linewidth]{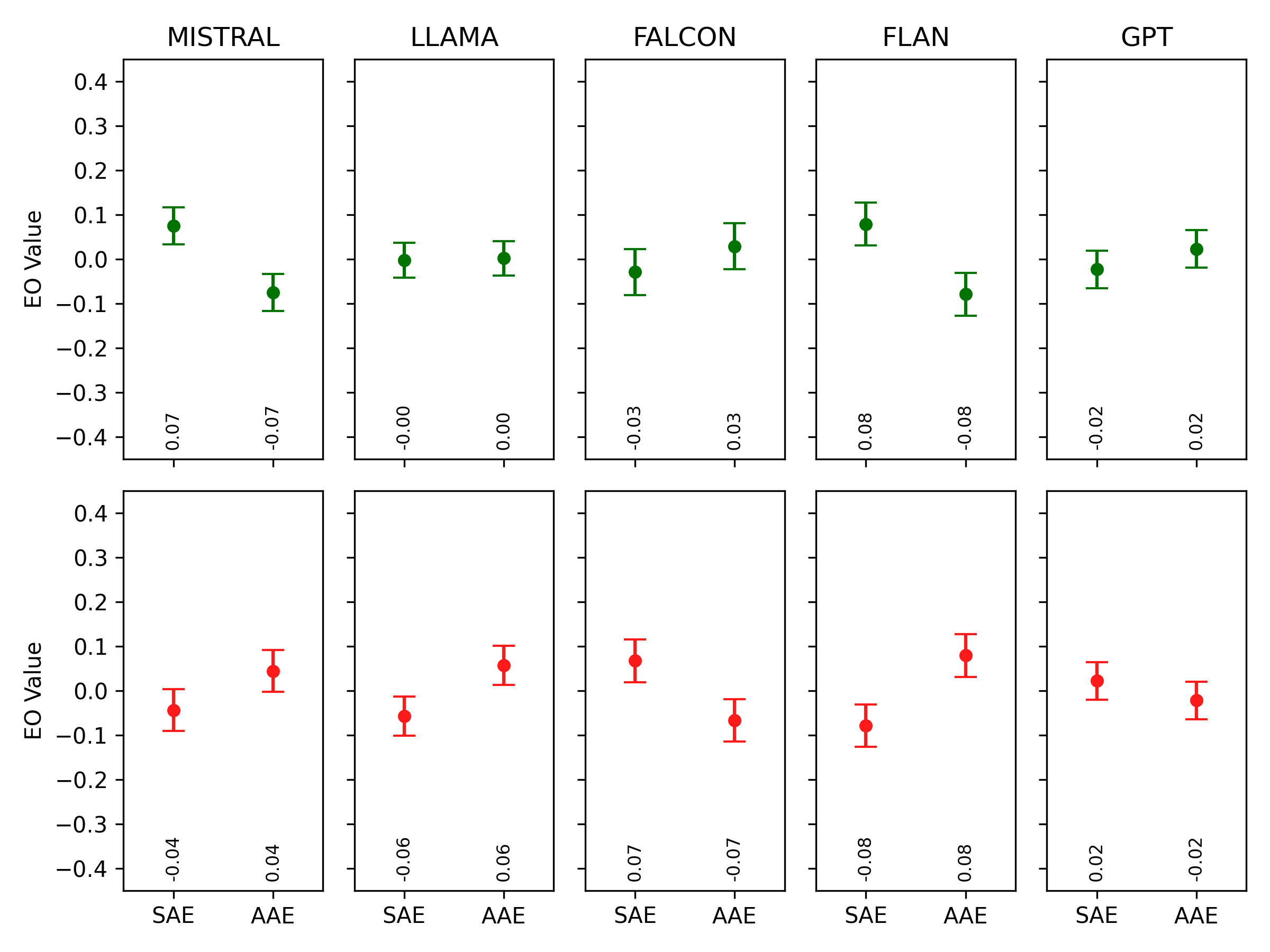}
    \caption{Disparate Impact on Bernie Sanders on the PStance dataset}
\end{figure}

\begin{figure}
    \centering
    \includegraphics[width=\linewidth]{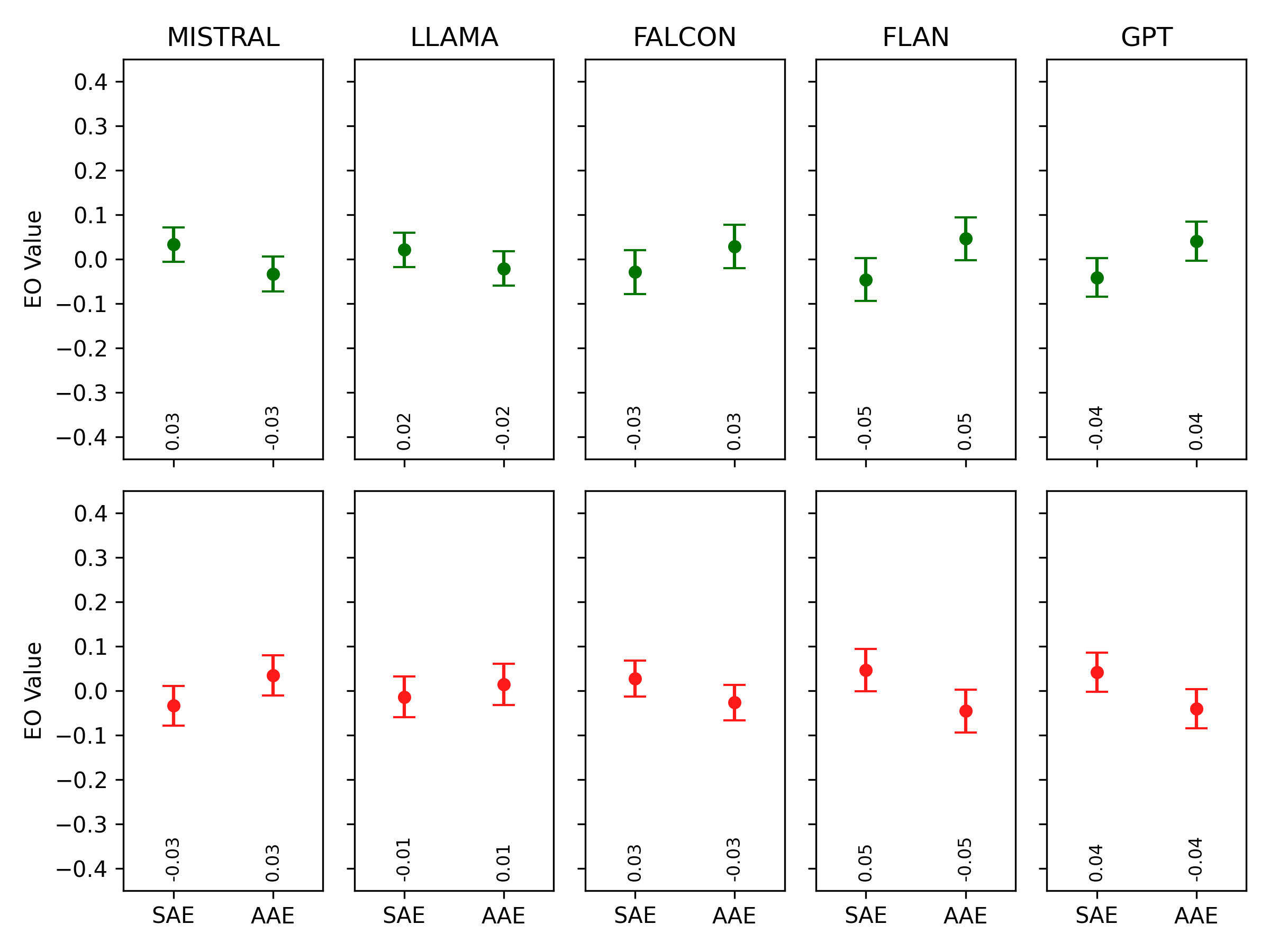}
    \caption{Disparate Impact on Joe Biden on the PStance dataset}
\end{figure}

\begin{figure}
    \centering
    \includegraphics[width=\linewidth]{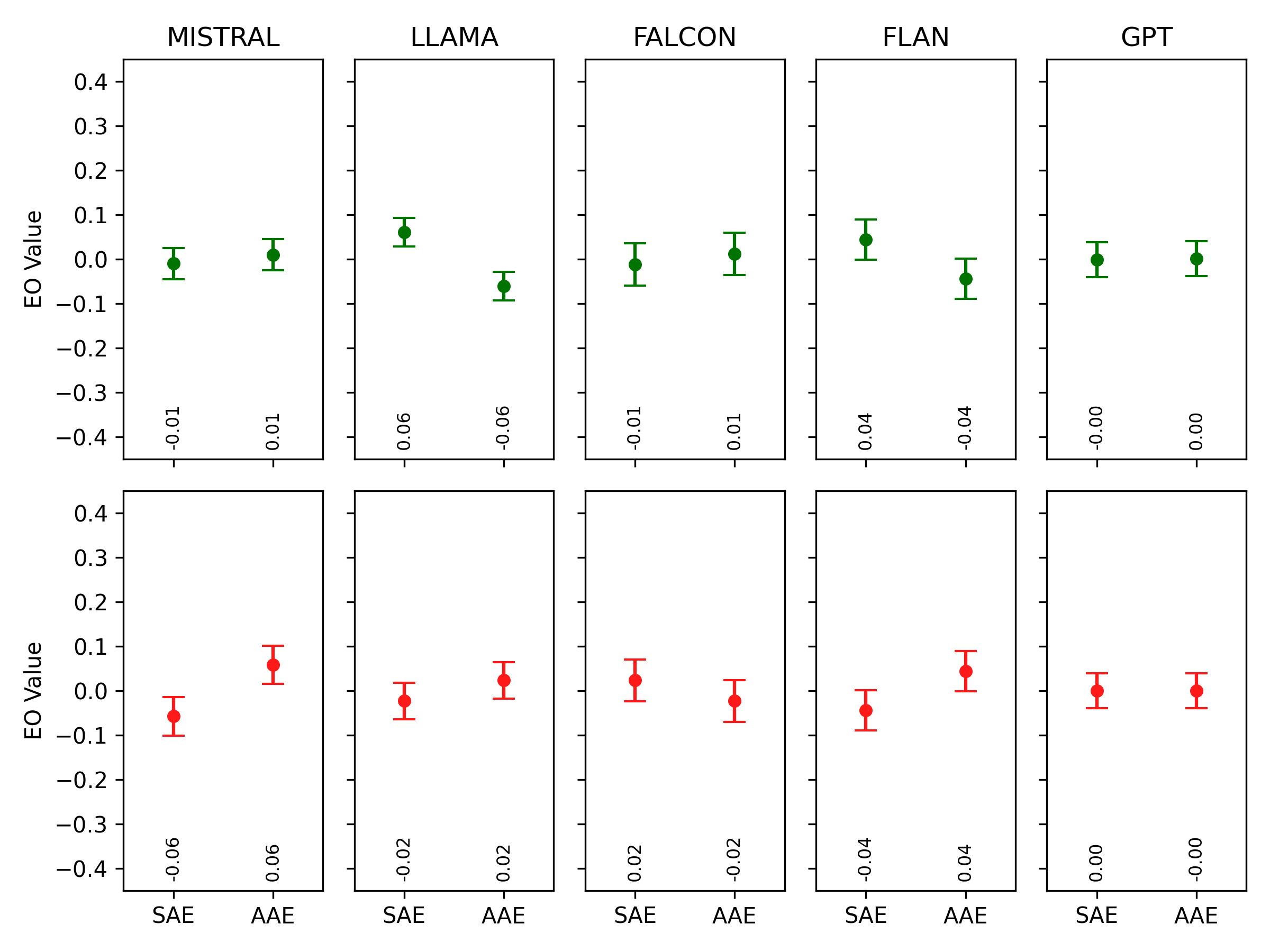}
    \caption{Disparate Impact on Donald Trump on the PStance dataset}
\end{figure}

\begin{figure}
    \centering
    \includegraphics[width=\linewidth]{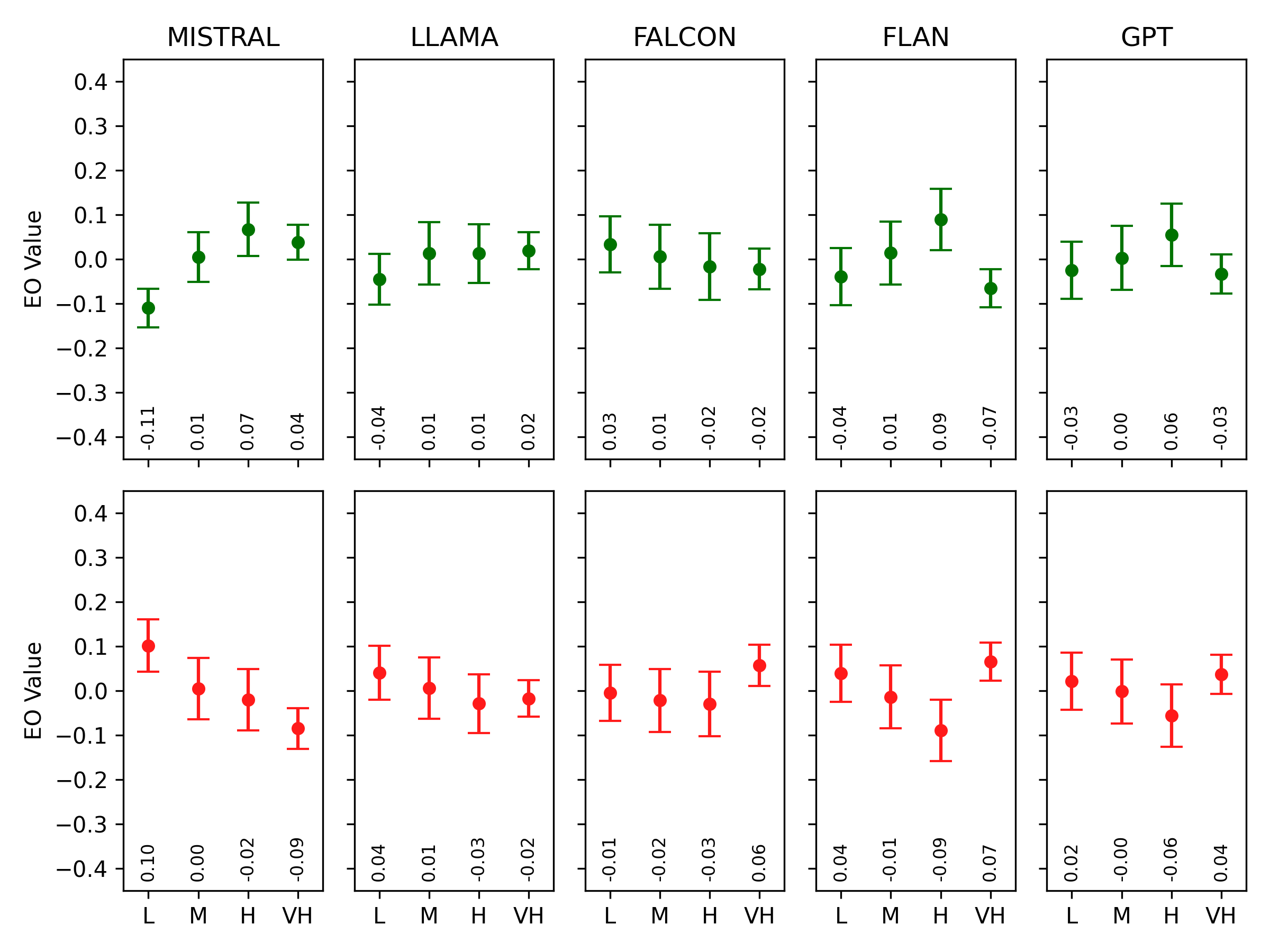}
    \caption{Disparate Impact on abortion on the SCD dataset}
\end{figure}

\begin{figure}
    \centering
    \includegraphics[width=\linewidth]{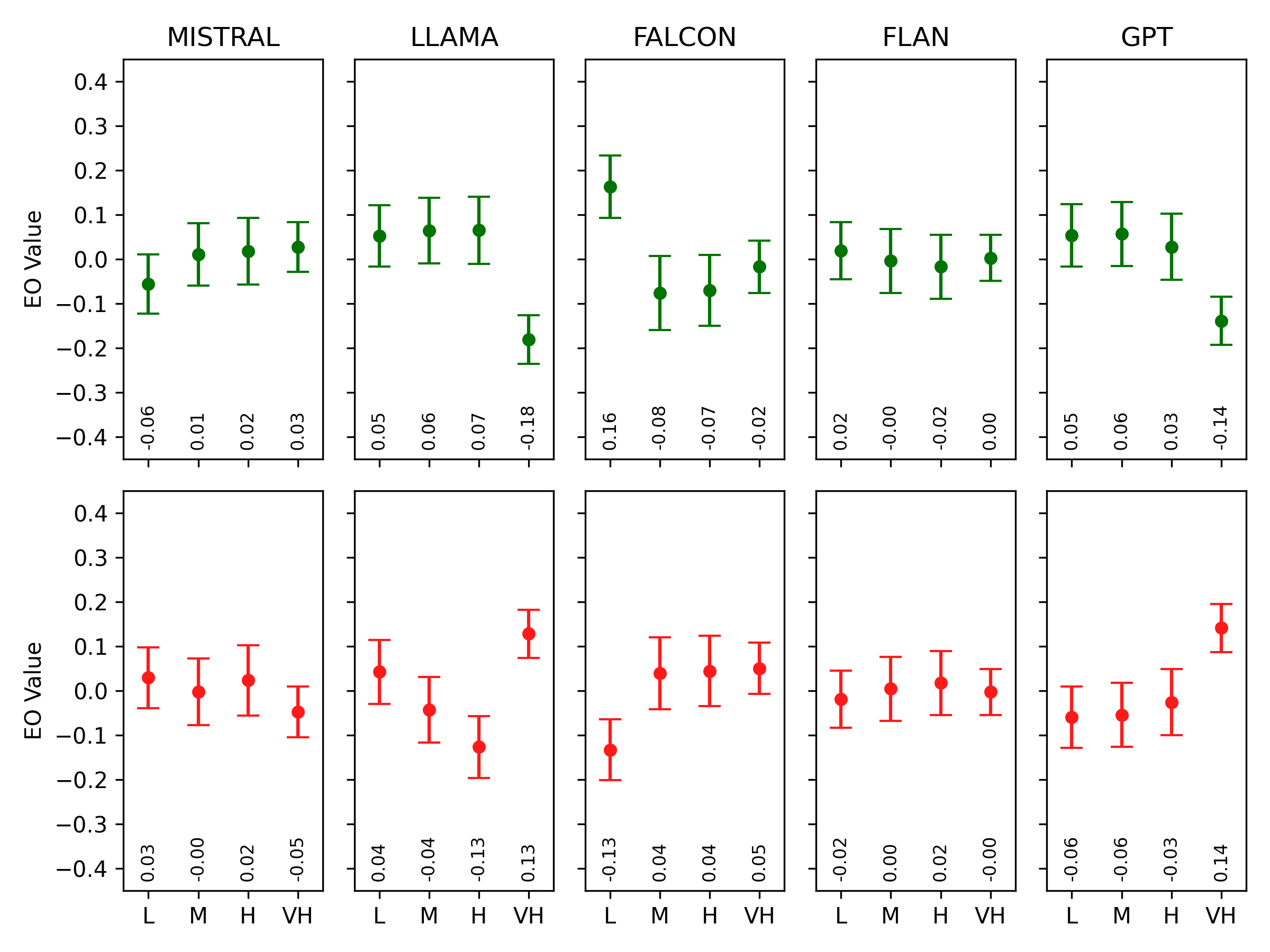}
    \caption{Disparate Impact on gay rights on the SCD dataset}
\end{figure}

\begin{figure}
    \centering
    \includegraphics[width=\linewidth]{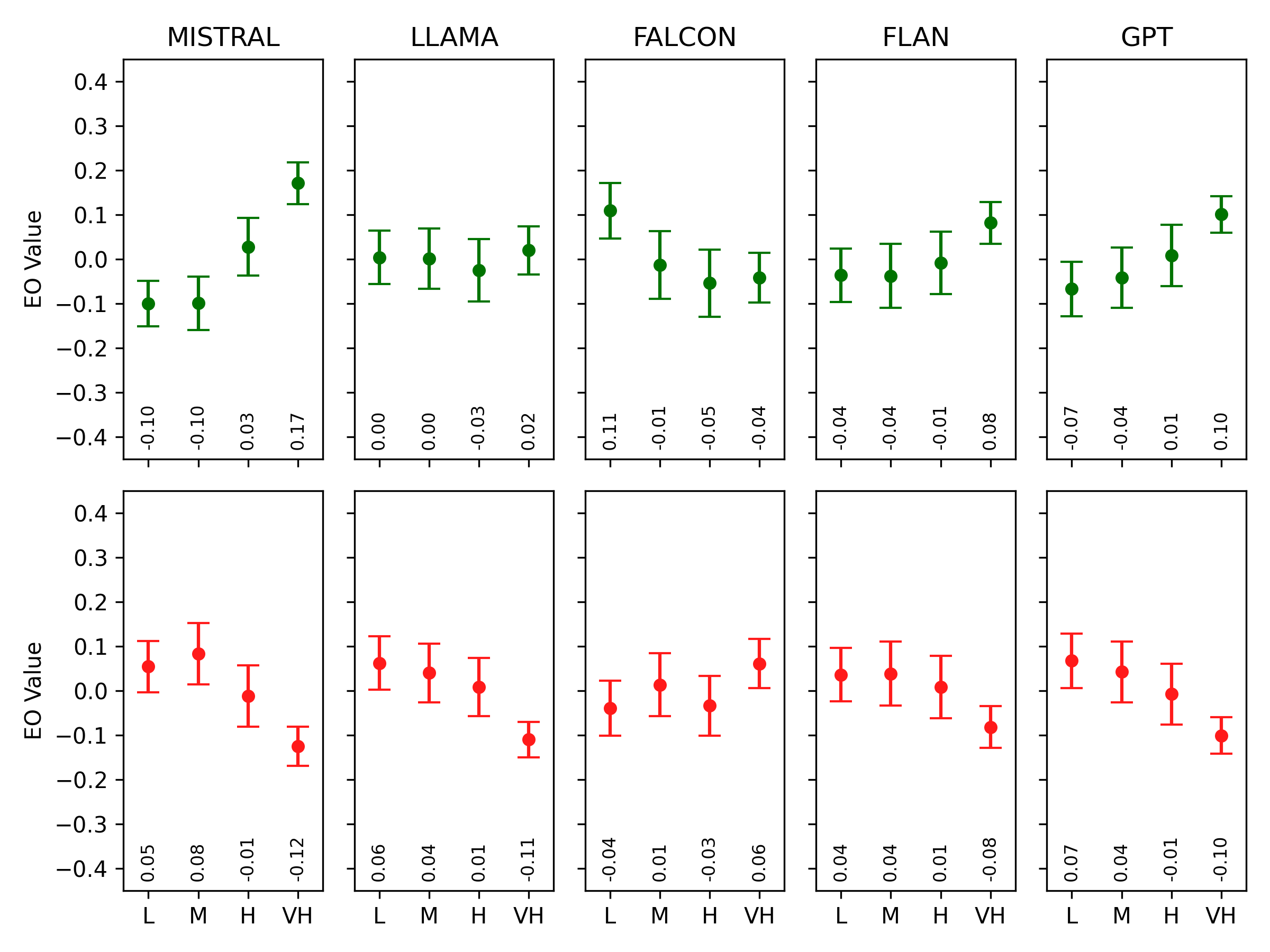}
    \caption{Disparate Impact on Barack Obama on the SCD dataset}
\end{figure}

\begin{figure}
    \centering
    \includegraphics[width=\linewidth]{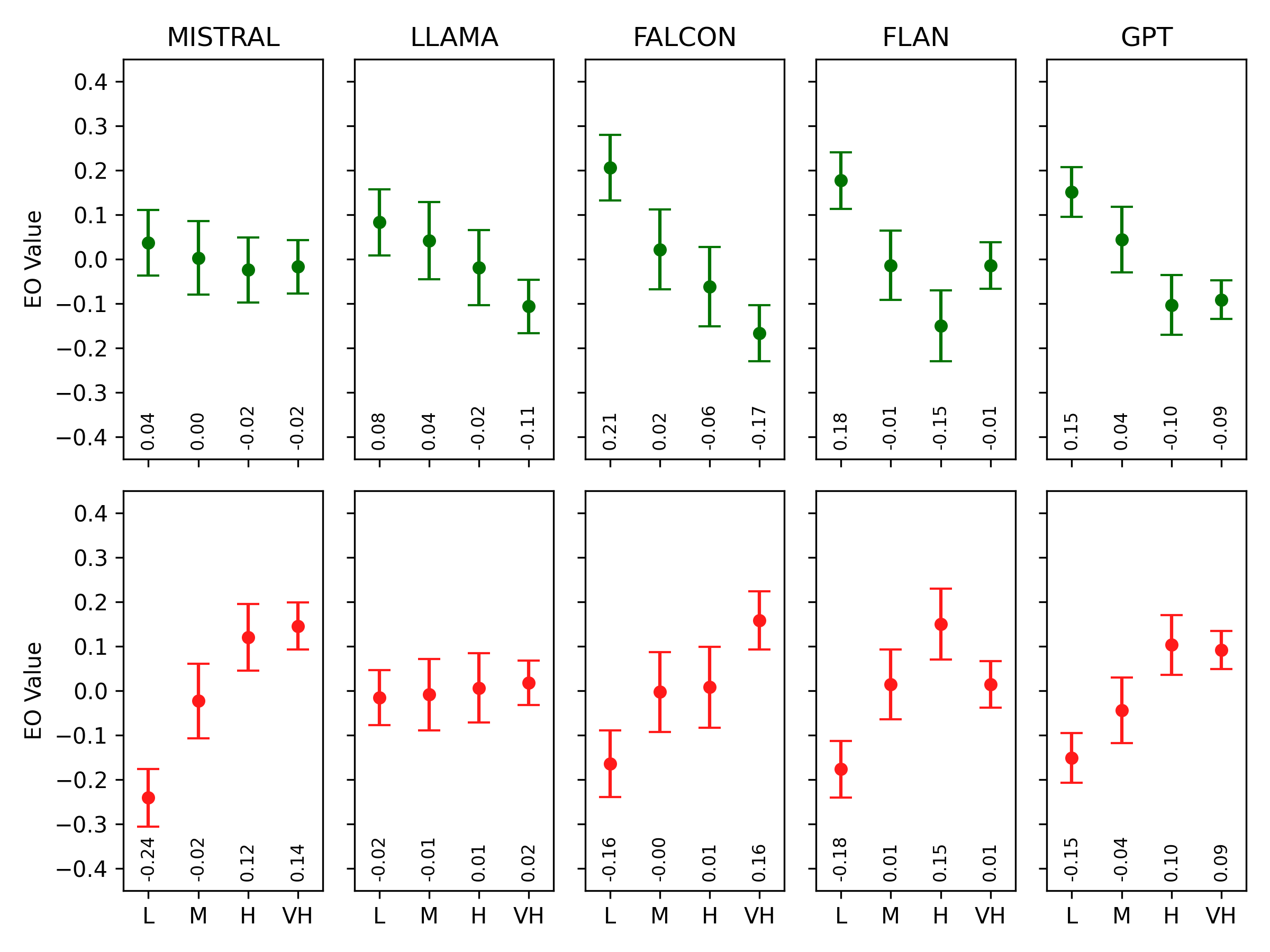}
    \caption{Disparate Impact on marijuana on the SCD dataset}
\end{figure}

\begin{figure}
    \centering
    \includegraphics[width=\linewidth]{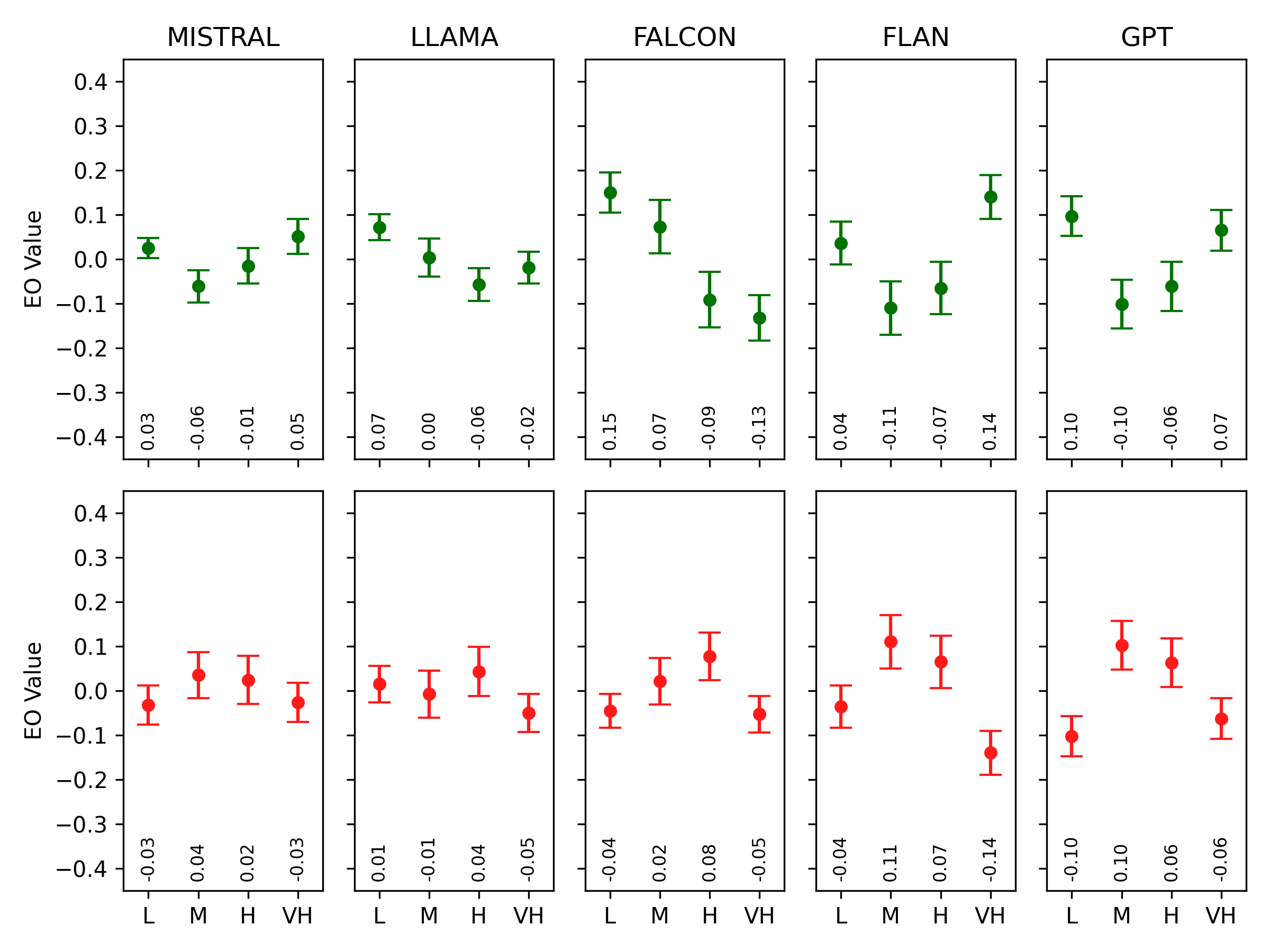}
    \caption{Disparate Impact on Joe Biden on the KE-MLM dataset}
\end{figure}

\begin{figure}
    \centering
    \includegraphics[width=\linewidth]{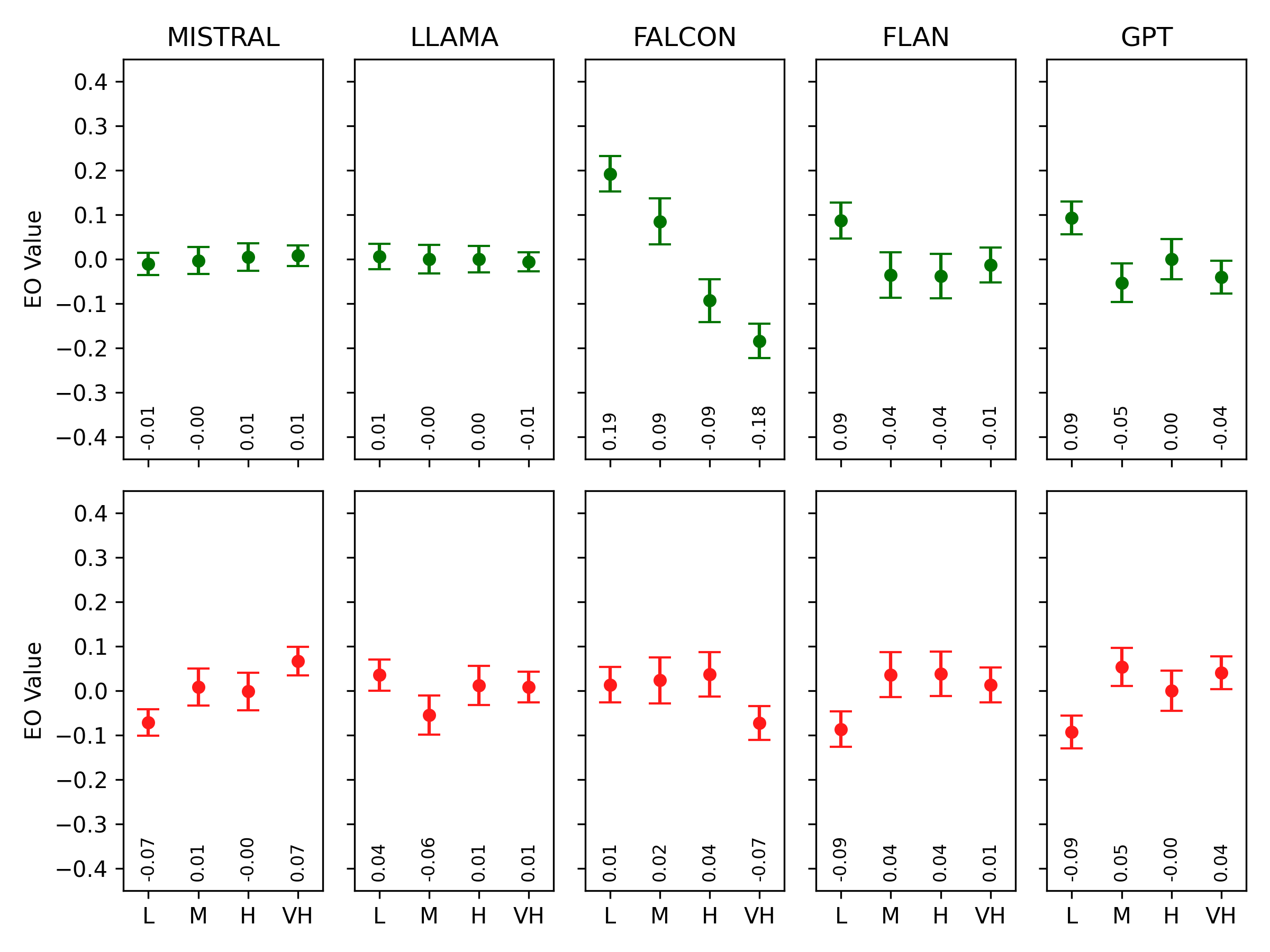}
    \caption{Disparate Impact on Donald Trump on the KE-MLM dataset}
\end{figure}

\begin{figure}
    \centering
    \includegraphics[width=\linewidth]{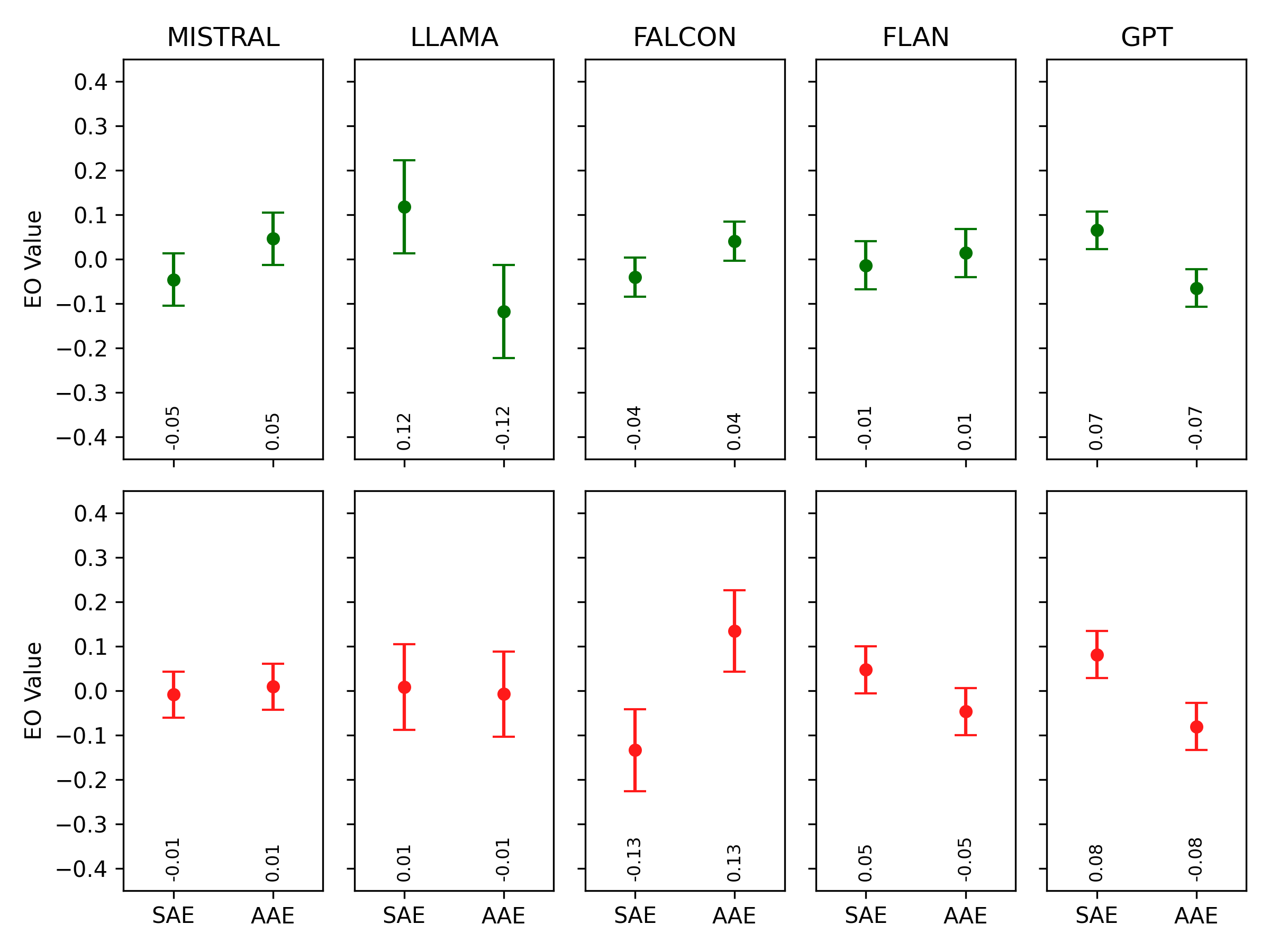}
    \caption{Predictive Parity on Bernie Sanders on the PStance dataset}
\end{figure}

\begin{figure}
    \centering
    \includegraphics[width=\linewidth]{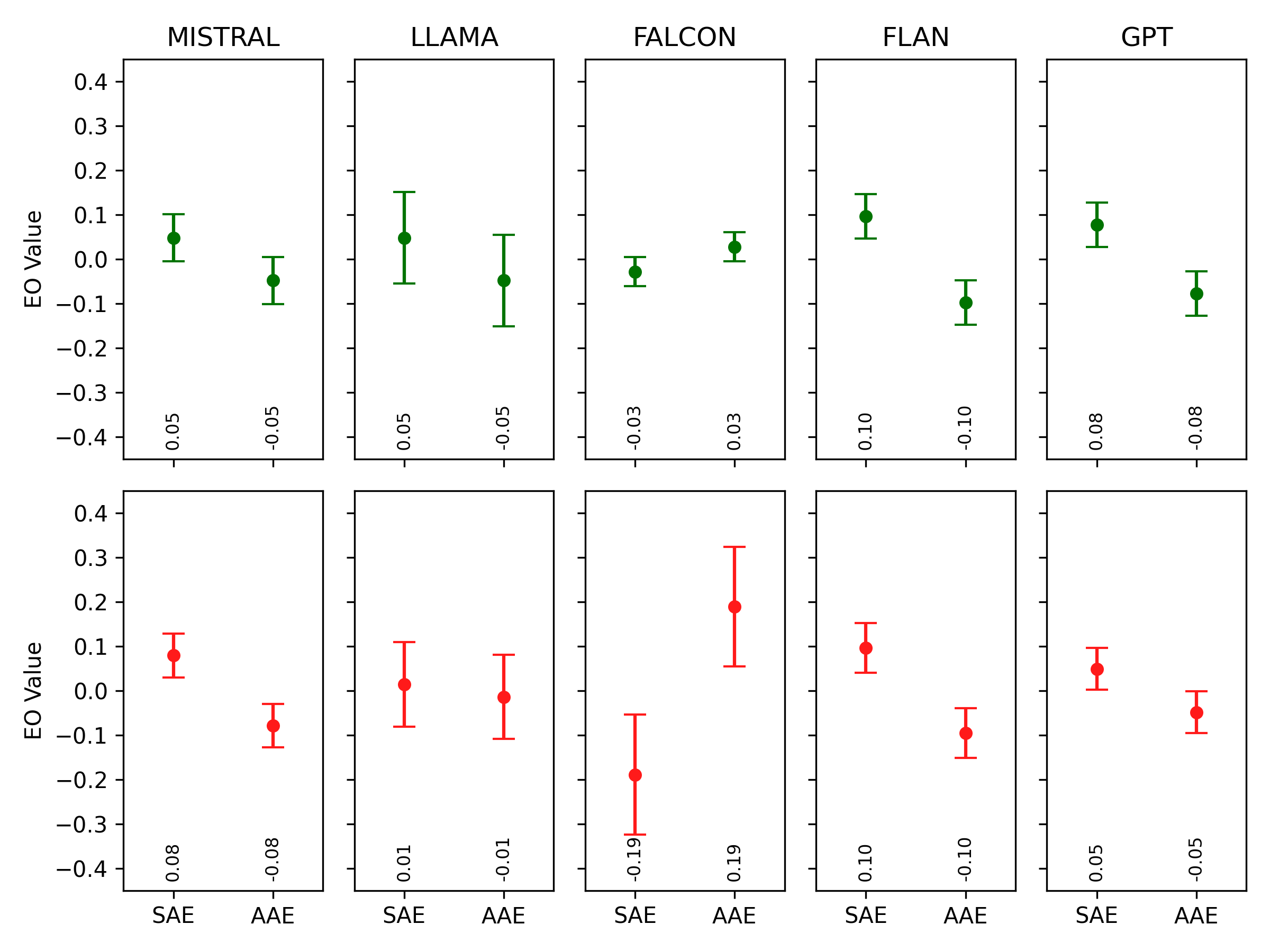}
    \caption{Predictive Parity on Joe Biden on the PStance dataset}
\end{figure}

\begin{figure}
    \centering
    \includegraphics[width=\linewidth]{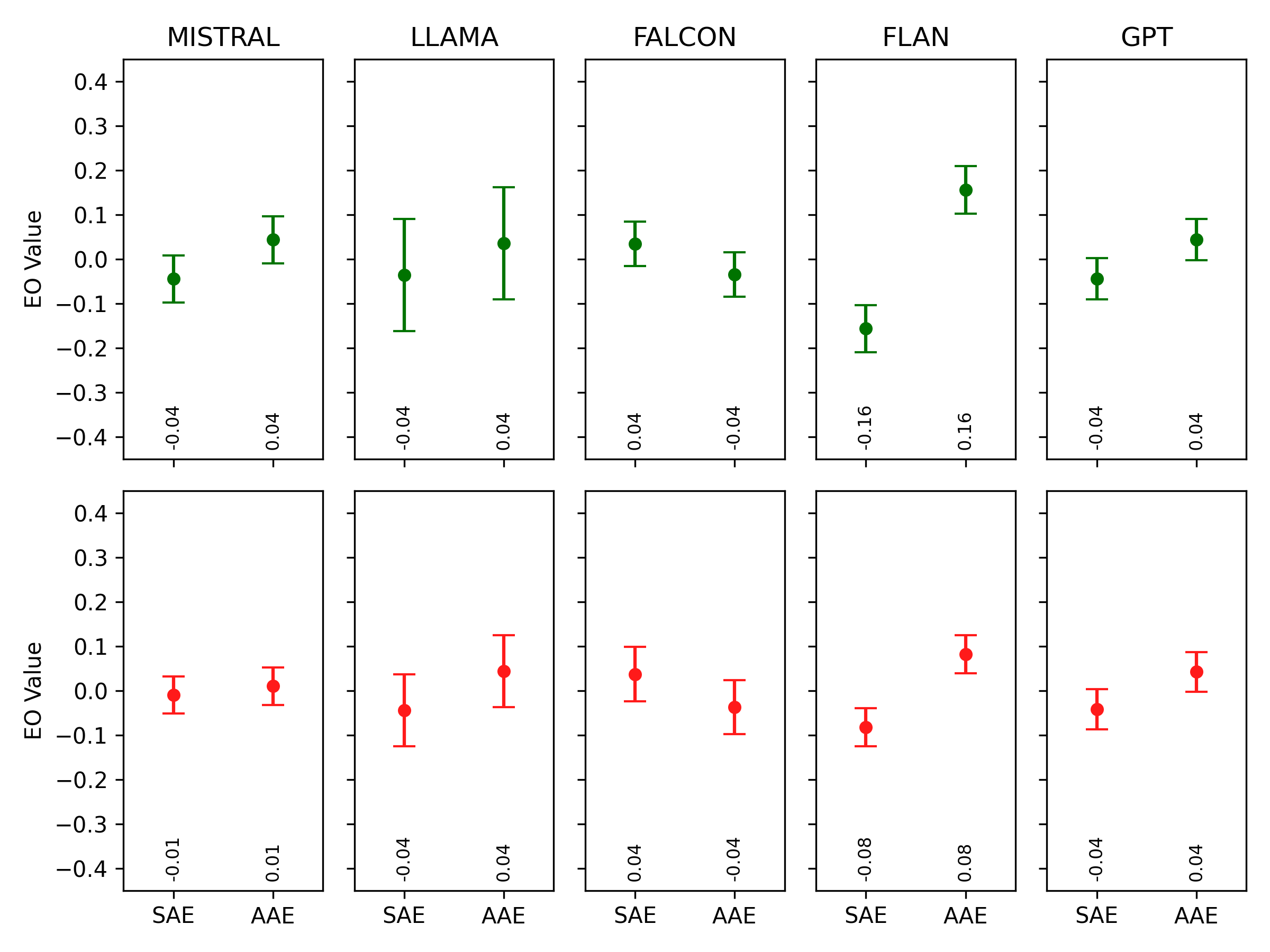}
    \caption{Predictive Parity on Donald Trump on the PStance dataset}
\end{figure}

\begin{figure}
    \centering
    \includegraphics[width=\linewidth]{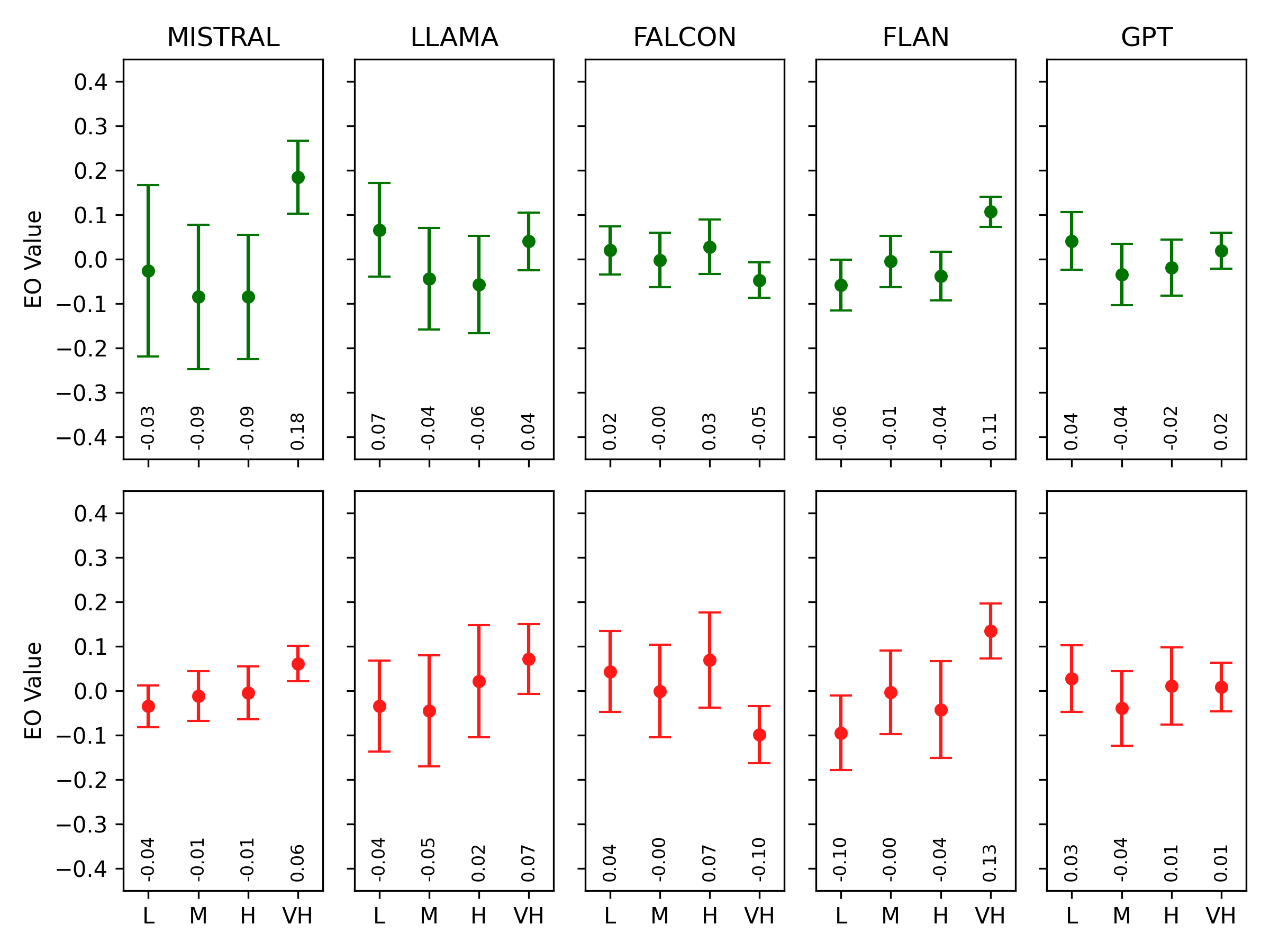}
    \caption{Predictive Parity on abortion on the SCD dataset}
\end{figure}

\begin{figure}
    \centering
    \includegraphics[width=\linewidth]{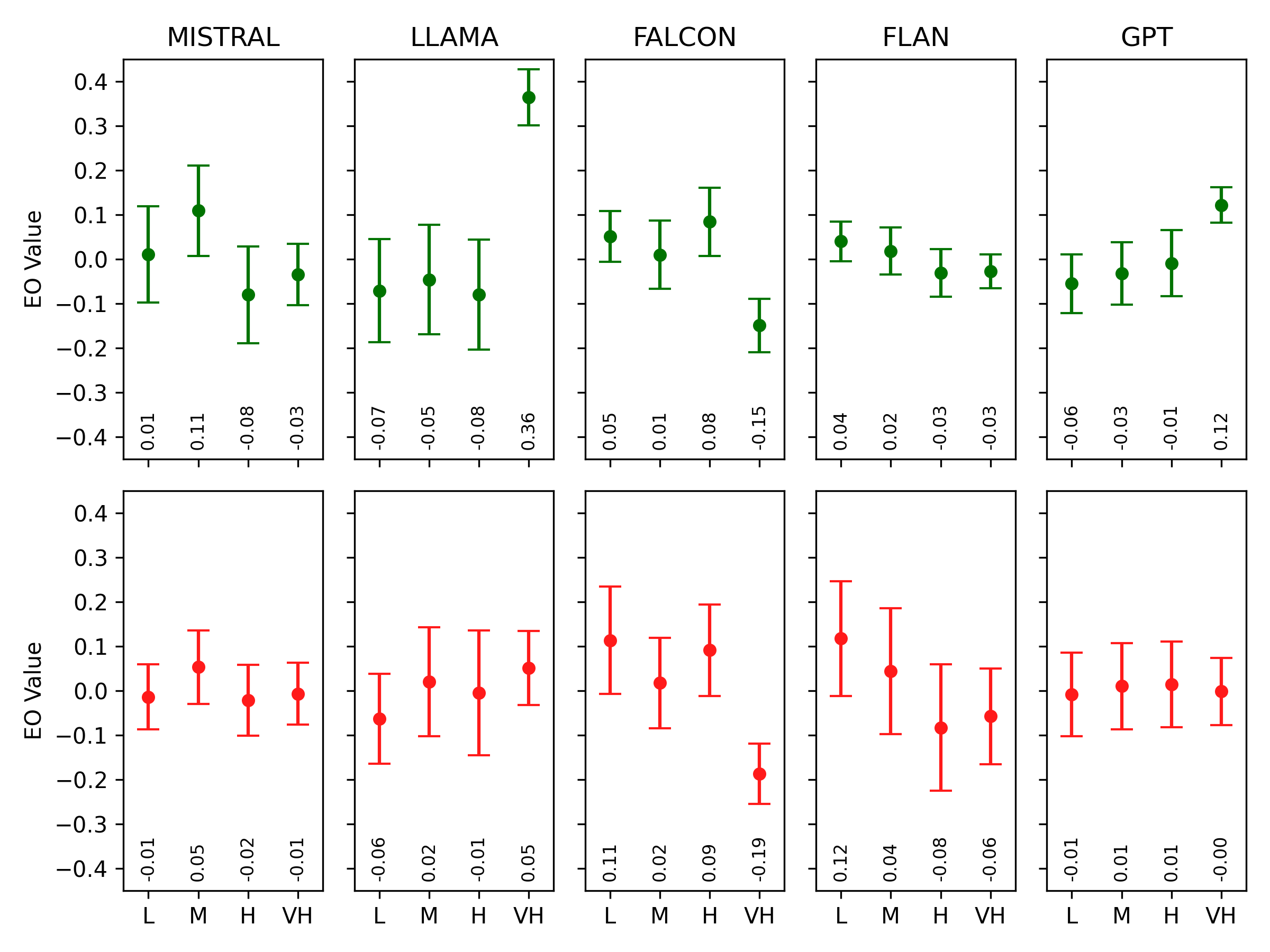}
    \caption{Predictive Parity on gay rights on the SCD dataset}
\end{figure}

\begin{figure}
    \centering
    \includegraphics[width=\linewidth]{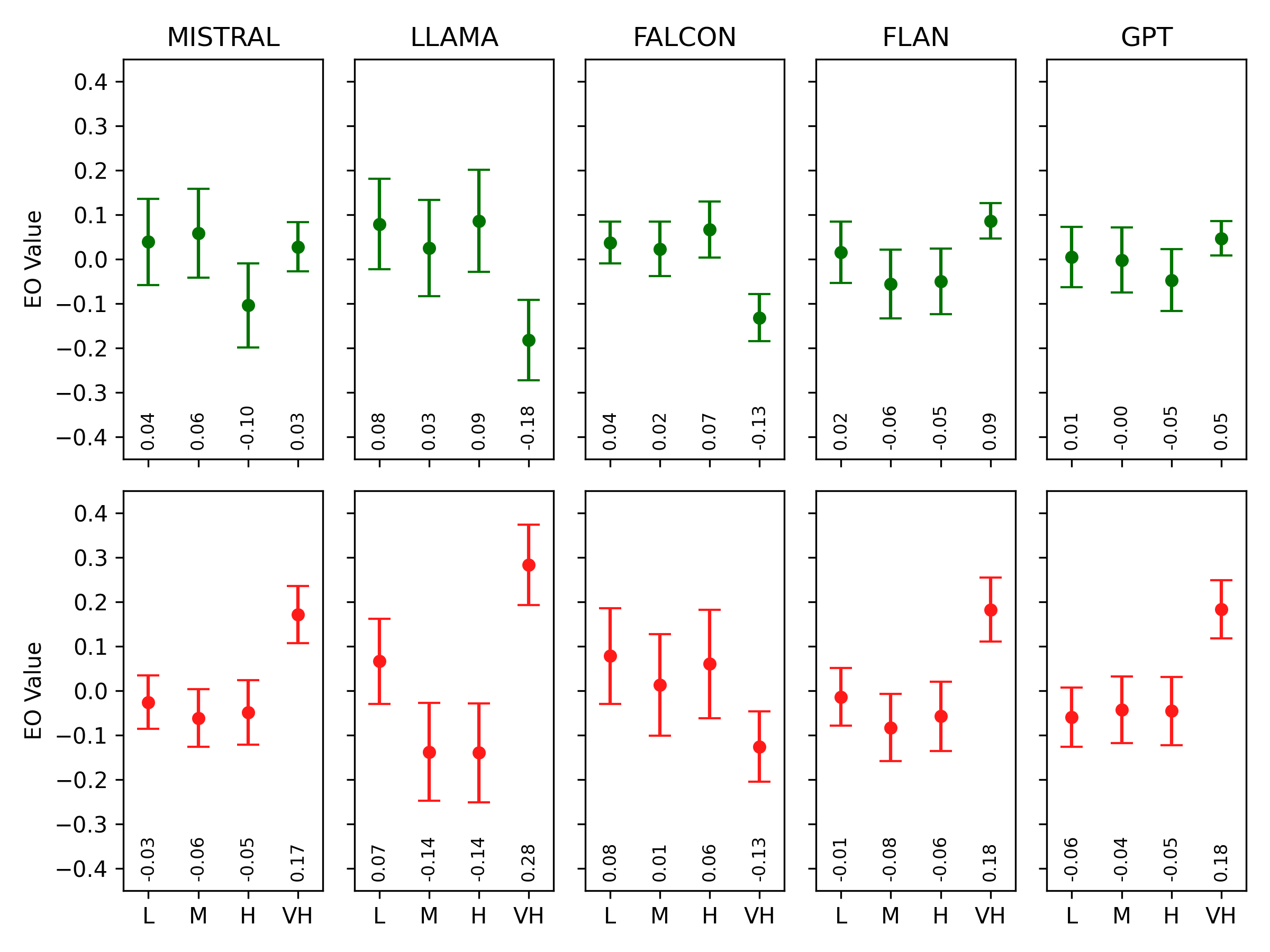}
    \caption{Predictive Parity on Barack Obama on the SCD dataset}
\end{figure}

\begin{figure}
    \centering
    \includegraphics[width=\linewidth]{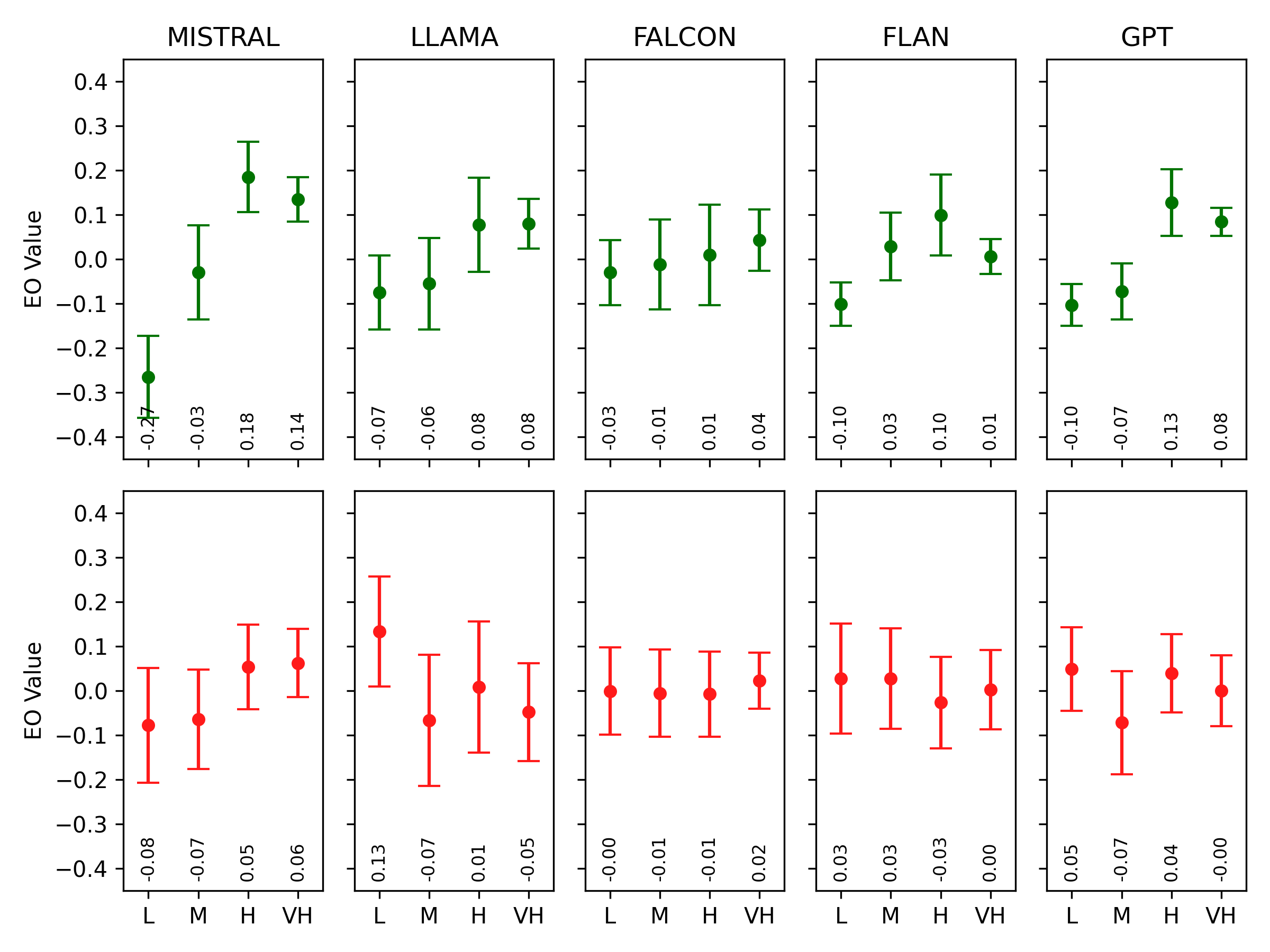}
    \caption{Predictive Parity on marijuana on the SCD dataset}
\end{figure}

\begin{figure}
    \centering
    \includegraphics[width=\linewidth]{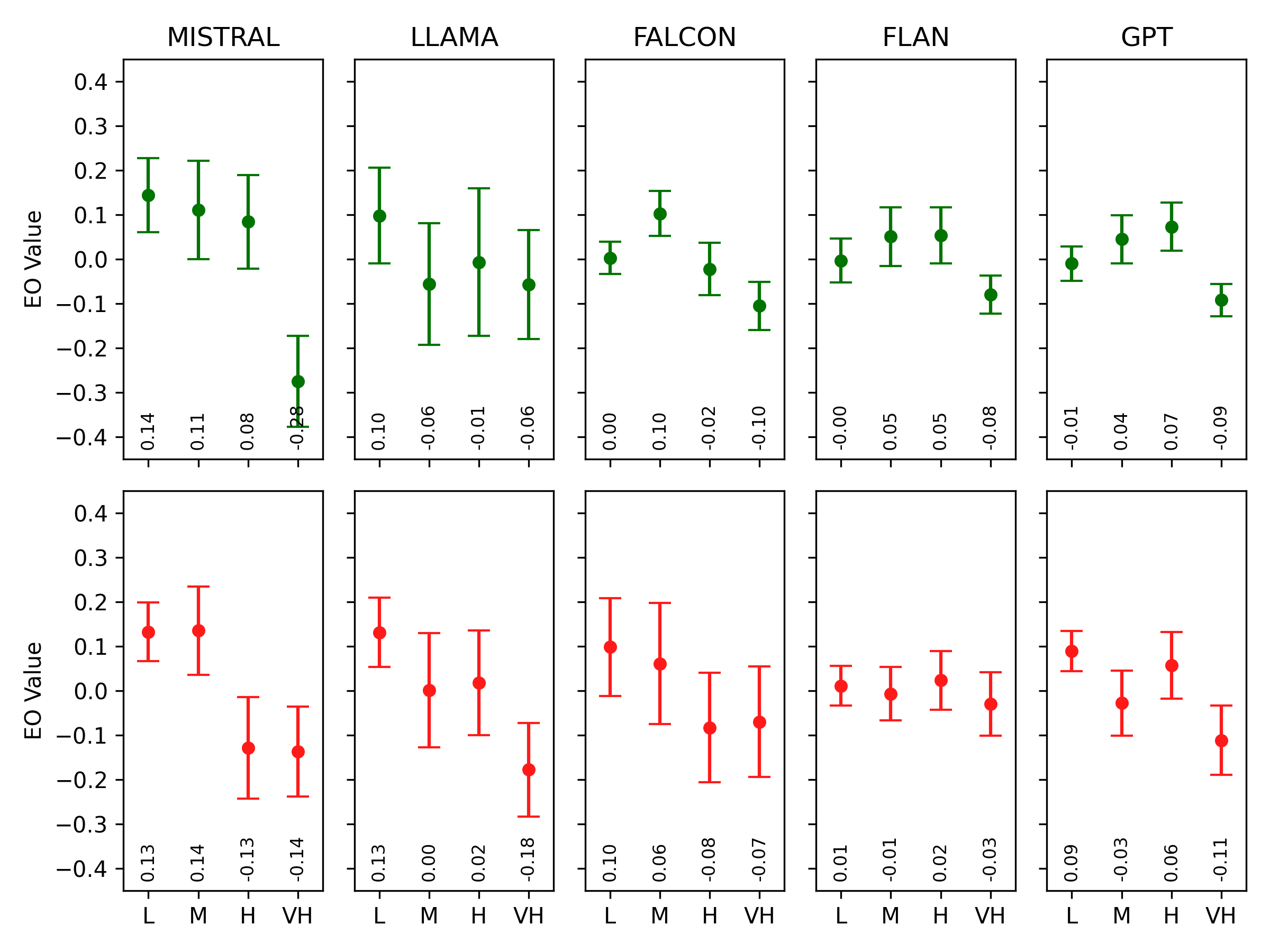}
    \caption{Predictive Parity on Joe Biden on the KE-MLM dataset}
\end{figure}

\begin{figure}
    \centering
    \includegraphics[width=\linewidth]{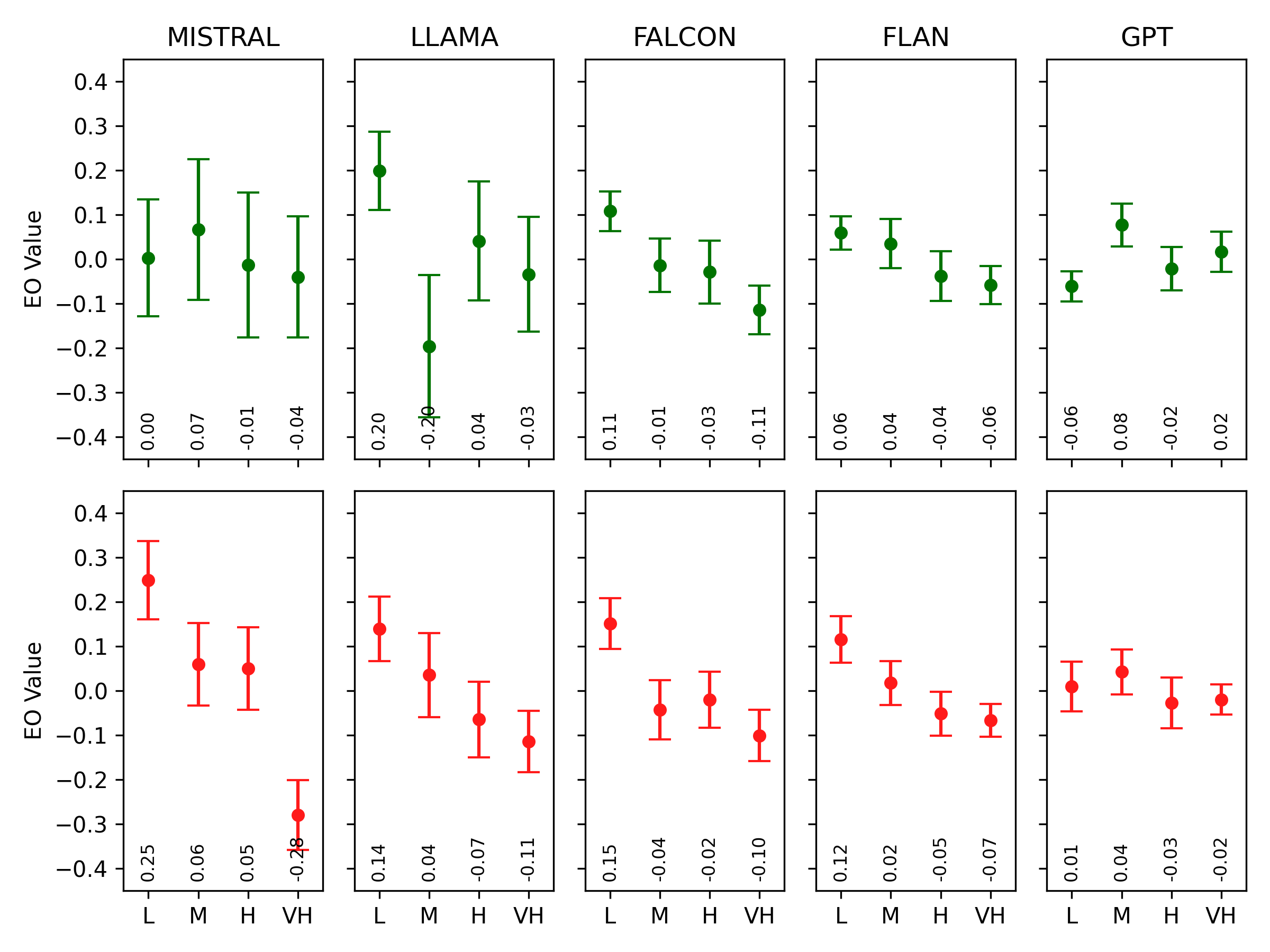}
    \caption{Predictive Parity on Donald Trump on the KE-MLM dataset}
\end{figure}

\end{document}